\documentclass[runningheads]{llncs}

\usepackage{eccv}

\usepackage{eccvabbrv}

\usepackage{graphicx}
\usepackage{booktabs}
\usepackage[dvipsnames,table]{}

\usepackage{enumitem}
\usepackage{multirow}
\usepackage{makecell}
\usepackage{stfloats}

\usepackage[accsupp]{axessibility}  

\usepackage{arydshln}

\newcommand{\myparagraph}[1]{\medskip\noindent\textbf{#1}}

\usepackage{hyperref}

\usepackage{orcidlink}
\usepackage[capitalize]{cleveref}
\crefname{section}{Sec.}{Secs.}
\Crefname{section}{Section}{Sections}
\Crefname{table}{Table}{Tables}
\crefname{table}{Tab.}{Tabs.}

\newcommand{\x}{\mathbf{x}}
\newcommand{\vit}{\mathcal{F}}
\newcommand{\lift}{\boldsymbol{\Theta}}

\begin{document}

\title{LiFT: A Surprisingly Simple Lightweight Feature Transform for Dense ViT Descriptors}
\titlerunning{LiFT: A Lightweight Feature Transform for Dense ViT Descriptors}

\def\thefootnote{*}\footnotetext{Equal contributors.}

\author{%
Saksham Suri\thefootnote{*}
\quad Matthew Walmer\thefootnote{*}
\quad Kamal Gupta
\quad Abhinav Shrivastava
}
\authorrunning{S.~Suri et al.}
\institute{University  of Maryland, College Park}

\maketitle

\begin{abstract}
We present a simple self-supervised method to enhance the performance of ViT features for dense downstream tasks. Our Lightweight Feature Transform (LiFT) is a straightforward and compact postprocessing network that can be applied to enhance the features of any pre-trained ViT backbone.
LiFT is fast and easy to train with a self-supervised objective, and it boosts the density of ViT features for minimal extra inference cost. Furthermore, we demonstrate that LiFT can be applied with approaches that use additional task-specific downstream modules, as we integrate LiFT with ViTDet for COCO detection and segmentation. Despite the simplicity of LiFT, we find that it is not simply learning a more complex version of bilinear interpolation. Instead, our LiFT training protocol leads to several desirable emergent properties that benefit ViT features in dense downstream tasks. This includes greater scale invariance for features, and better object boundary maps. By simply training LiFT for a few epochs, we show improved performance on keypoint correspondence, detection, segmentation, and object discovery tasks. Overall, LiFT provides an easy way to unlock the benefits of denser feature arrays for a fraction of the computational cost. For more details, refer to our \href{https://www.cs.umd.edu/~sakshams/LiFT/}{\textcolor{magenta}{project page}}.

\end{abstract}

\section{Introduction}

\begin{figure}[h]
    \centering
    \begin{subfigure}{\linewidth}
        \centering
        \includegraphics[width=\linewidth]{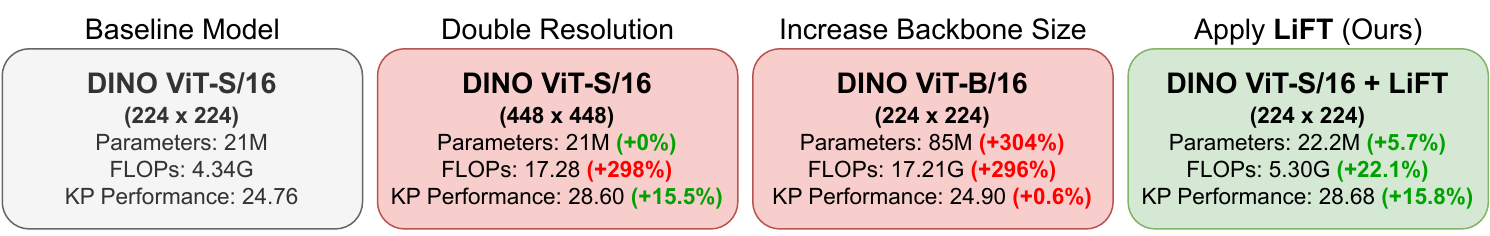}
    \end{subfigure}
    \hrule
    \begin{subfigure}{\linewidth}
        \centering
        \includegraphics[width=\linewidth]{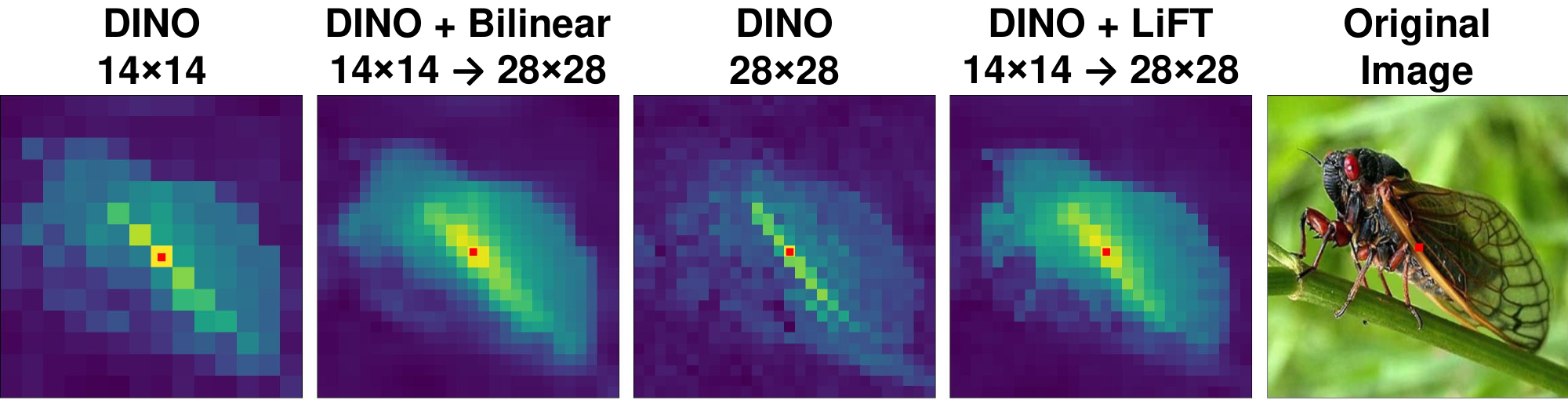}  
    \end{subfigure}
    \caption{\textbf{(Top)} Increasing the backbone size or doubling the input resolution can boost the effectiveness of self-supervised ViT features for dense tasks like keypoint (KP) correspondence. However, both of these options come at a significant cost in terms of parameter count, inference cost, or both. We present \textbf{LiFT}, a surprisingly simple \textbf{Li}ghtweight \textbf{F}eature \textbf{T}ransform that unlocks the benefits of dense self-supervised ViT representations for minimal extra cost. \textbf{(Bottom)} LiFT also has useful emergent properties, such as yielding cleaner object boundaries in feature similarity maps.}
    \label{fig:teaser}
\end{figure}

In recent years, Vision Transformers (ViTs)~\cite{dosovitskiy2020vit} have emerged as preferred architectures for many image and video recognition tasks in the Computer Vision community. They also represent a major design shift compared with the well-explored Convolutional Neural Networks (CNNs). ViTs typically convert images into a very coarse grid of image patches (or tokens) before applying transformer layers. This allows ViTs to learn increasingly powerful patch-wise representations in successive layers~\cite{walmer2022teaching}. The expressive power of ViTs stems from their wide receptive field throughout all layers made possible by multi-headed self-attention operations~\cite{vaswani2017attention}. The downside of this design is that despite being able to learn powerful representations, ViTs often lack spatial granularity in their features due to the low resolution of the token/patch grid. This hinders their off-the-shelf application to dense and local tasks such as object detection, segmentation, and keypoint correspondence.
Increasing the feature resolution of a ViT directly using a larger image size or smaller patch size leads to an increased number of patches. Self-attention, being a quadratic operation, grows in memory consumption as $\mathcal{O}(N^2)$ where $N$ is the number of patches in the image.
Prior works have proposed alterations to the ViT architecture to make it better suited for dense tasks, but their methods either involve expensive carefully designed training, task-specific loss functions or heuristics, or high inference costs~\cite{Yun_2022_CVPR, ziegler2022self, amir2021deep}.

In this work, we propose a simple \textbf{Li}ghtweight \textbf{F}eature \textbf{T}ransform or \textbf{LiFT} to generate dense ViT features that provide significant performance gains in downstream tasks such as detection, segmentation, keypoint correspondence, and object discovery.
LiFT can unlock the benefits of dense feature representations for a fraction of the computational cost compared with other approaches. 
As illustrated in Figure~\ref{fig:arch},
our proposed method fuses the coarse high-level information of ViT features with convolution-based image features derived from the original image to generate higher-density feature maps without incurring the high computational cost of extra tokens.
We show that this approach does not require any complex training recipe and, once trained on a general purpose dataset, generalizes well to multiple downstream tasks. Our approach can be trained with a simple self-supervised loss and generalizes to input image resolutions not seen during training. LiFT can be readily plugged on top of any ViT backbone to enhance its features, 
and it can also be integrated into pipelines that use additional task-specific downstream modules, like the Mask-RCNN head~\cite{he2017mask} used by ViTDet~\cite{li2022exploring}.
Additionally, we show that we can apply LiFT in a recursive manner to increase feature resolution even further.

We demonstrate the effectiveness of LiFT quantitatively on `local' tasks, which require features computed at precise locations, as well as on `dense' tasks, which require features computed for the entire image.
Specifically, we present results for LiFT applied to SPair-71k Keypoint Correspondence~\cite{min2019spair},
COCO Detection and Segmentation~\cite{lin2014microsoft}, DAVIS Video Segmentation~\cite{pont20172017}, and Unsupervised Object Discovery on Pascal VOC 2007~\cite{pascal-voc-2007}, Pascal VOC 2012~\cite{pascal-voc-2012} and COCO20K~\cite{Lin2014MicrosoftCC}. 
For all of these tasks, LiFT is able to meet or exceed the performance of prior works for a fraction of the computational cost.
As an example, in \Cref{fig:teaser} we compare three options for boosting performance in SPair-71k Keypoint Correspondence. LiFT provides a significant performance gain while increasing the total parameter count of the network by a mere $5.7\%$. This is compared to the $304\%$ parameter count increase incurred by the step up from ViT-S/16 to ViT-B/16. Increasing the input resolution is a trivially easy way to boost the feature density, and it also gives improved performance. However, it increases the total inference FLOPs by almost $300\%$ while LiFT only increases the cost by $22.1\%$, giving a far superior compute cost \vs performance trade off.

Despite the simplicity of our LiFT approach, we show that it is not just learning a more complex version of bilinear upsampling. Instead, we demonstrate that LiFT has several desirable emergent properties that enhance ViT features to make them better suited for dense tasks. We find that LiFT improves the scale invariance of ViT features, as measured using Centered Kernel Alignment (CKA)~\cite{Cortes2012AlgorithmsFL, kornblith2019similarity}. We also qualitatively show that LiFT yields better object boundary maps when computing feature similarity maps. Overall, LiFT represents an orthogonal avenue of improvement compared to prior dense ViT feature extraction strategies, and it can be combined with past methods to further advance self-supervised performance on dense prediction tasks. In summary, our contributions are as follows:
\begin{itemize}[leftmargin=*,itemsep=0em]
    \item We propose \textbf{LiFT}, a \textbf{Li}ghtweight \textbf{F}eature \textbf{T}ransform that boosts the performance of existing ViT features on dense and local downstream tasks using a simple, quick training and inference strategy.
    \item We show that LiFT boosts the performance of self-supervised ViT features for detection, segmentation, keypoint correspondence, and object discovery tasks.
    \item We demonstrate the adaptability of LiFT for any ViT backbone by showing improvements with DINO~\cite{caron2021emerging}, MoCo~\cite{chen2021mocov3} and Supervised ViT features. Additionally, LiFT even works on image resolutions not used during training.
    \item We show that LiFT features have desirable emergent properties like improved scale invariance and better feature alignment with object boundaries.
\end{itemize}

\section{Related Work}

\subsection{Vision Transformers}
Recently, Vision Transformers (ViTs)~\cite{dosovitskiy2020vit} have gained wide popularity as general-purpose models for multiple computer vision tasks such as image classification~\cite{dosovitskiy2020vit,touvron2021training,ali2021xcit}, object detection~\cite{zhu2020deformable, li2022exploring, liu2021swin}, segmentation~\cite{ranftl2021vision,strudel2021segmenter}, video classification~\cite{arnab2021vivit, bertasius2021space}, and more. Many of these methods adapt ViTs to different tasks by making suitable changes to the output heads \cite{li2022exploring, xu2022vitpose}. There have also been multiple variants of ViT like Swin~\cite{liu2021swin}, MViT~\cite{fan2021multiscale}, and PVT~\cite{wang2021pyramid} which incorporate hierarchy and multiscale learning. Additionally, there are some works which try to bridge the gap between transformers and CNNs by incorporating convolutions into ViT architectures~\cite{wu2021cvt,d2021convit}. In this work, we focus on improving traditional or ``Plain'' ViT backbones, as they are the most general and widely adopted form of ViT. These models learn powerful representations, but they suffer in terms of feature resolution, an issue which we aim to address with LiFT.
Many self-supervised ViT pretraining methods have been proposed to learn powerful off-the-shelf features, such as DINO~\cite{caron2021emerging}, MoCo~\cite{chen2020improved,he2020momentum, chen2021empirical}, and MAE~\cite{he2022masked}.
Prior works~\cite{walmer2022teaching, ghiasi2022vision, shekhar2023objectives} have shown that different pretraining leads to significant differences in the properties of the learned features. We show that LiFT can improve the quality and usefulness of ViT features for a range of pretraining methods.

\subsection{Extracting Denser Features from ViTs}
Works like~\cite{amir2021deep} and~\cite{walmer2022teaching} show the general benefits of denser feature maps for local tasks, and multiple strategies have been proposed to extract denser feature arrays from pretrained ViTs.
ViT-Adapter~\cite{chen2022vision} applies ViTs to dense tasks through the use of finetunable side-networks for feature pyramid extraction. Like ViT-Adapter, LiFT aims to enhance ``Plain'' ViT features for dense tasks, however, the ViT-Adapter method is not task-agnostic, and it is trained in a fully supervised way with full detection and segmentation labels for the downstream dataset. In contrast, LiFT is a task-agnostic general enhancement for ViT features, and it is trained with a completely self-supervised objective.
LiFT is also faster and easier to train, as it does not require passing gradients through the ViT backbone, and it is also ${\sim}4.8\times$ smaller than the similar ViT-Adapter. ~\cite{amir2021deep} proposes a simpler technique by reducing the stride during initial image patch extraction. This does not require any training, but also becomes computationally expensive as the number of tokens increase quadratically requiring more GPU memory and FLOPs. Our LiFT approach can be thought of as a ``shortcut'' to achieve the benefits of denser features for a fraction of the cost of increased tokens.

\subsection{Feature Upsampling}
Over the years, many works have been proposed to convert lower resolution features into higher resolution ones. This includes basic methods like bilinear interpolation, and Joint Bilinear Upsampling (JBU) \cite{kopf2007joint}, which leverages high-frequency information from the input image to improve bilinear interpolation. Our LiFT module leverages the input image similarly. Resize-convolutions \cite{odena2016deconvolution} are proposed as an alternative to deconvolution. Many more sophisticated upsampling operators have been proposed~\cite{wang2019carafe, lu2022sapa, lu2022fade, lu2020index, dai2021learning}, however these methods are designed to be trained in the context of their particular architecture and downstream task, unlike our LiFT module, which is stand-alone and task-agnostic. Moreover, many of these methods are designed to be integrated with convolutional encoder/decoder architectures, and thus are not easily applied to ViT backbones.
\cite{tan2018feature} proposes a GAN-based approach for CNN feature densification which requires careful training and a mixture of adversarial and focal loss.
In comparison, our LiFT approach is easy to train through a simple self-supervised objective.
Other works, like~\cite{chen2022super} and~\cite{zhu2019low}, have applied student/teacher distillation with feature super-resolution to improve CNN classification performance on low-resolution images. While we also perform feature super-resolution, the aim of our work is not to distill a student network, but rather we aim to generate densified features while avoiding finetuning the ViT backbone.

Finally, we would like to acknowledge a concurrently published work, FeatUp~\cite{fufeatup}, which presents two feature upsampling methods, one based on JBU and a second based on implicit network learning.
Like LiFT, FeatUp trains general-purpose stand-alone feature upsamplers for multiple backbones and downstream tasks, though its learning process is more complex, as it requires training an additional downsampler module which is later discarded. LiFT and FeatUp$_{\text{(JBU)}}$ are both feedforward, making them easy to apply to any image, though we believe LiFT's feature maps are sharper and better matched to the object boundaries than the JBU-enhanced features. FeatUp$_{\text{(Implicit)}}$ produces sharp, impressive feature maps, though compared with LiFT and FeatUp$_{\text{(JBU)}}$, it is a much less scalable method, as it trains a new implicit network for every input image. Overall, we believe LiFT captures the best properties of both FeatUp variations.

\subsection{Finetuning ViTs for Dense Tasks}
Another line of research has focused on finetuning self-supervised ViT backbones to make their features better suited to dense, locally-focused tasks.
SelfPatch~\cite{ziegler2022self} and Leopart~\cite{Yun_2022_CVPR} both use pretrained DINO models as a starting point and improve their patch-level representations using dense self-supervised tasks.
SelfPatch learns by enforcing similarity to neighboring patch features, and Leopart uses spatially dense clustering to enforce similarity within clusters to learn better part representations. These approaches finetune the full backbone with specialized training strategies and losses, which can still be expensive to train and not easily extendable to other backbones. In contrast, our approach does not require backbone finetuning at all, and instead trains a lightweight post-processing module. LiFT
can be easily trained on ImageNet~\cite{deng2009imagenet} in a self-supervised manner in very few epochs and afterwards it can generalize to multiple downstream tasks.
\section{Method}

\subsection{ViT Background}

Consider a ViT that takes an image with dimensions $H \times W \times 3$ as input and outputs feature descriptors at the resolution $\frac{H}{P} \times \frac{W}{P}$, where $P$ is the patch size. $P$ is usually 8 or 16 depending on the ViT variant. We typically assume that the patch-extraction stride length, $S$, is equal to $P$, though this can be altered, as proposed by \cite{amir2021deep}. For now we will assume that $S=P$.
Our goal is to transform the coarse, low-resolution features of a pretrained ViT into dense feature descriptors without having to re-train or finetune the ViT. A na\"ive way to achieve this transformation could be to scale up the input image to  $CH \times CW$ resolution to achieve a feature grid that is $C$ times larger along both dimensions. This approach can result in a significant increase in memory consumption and can be prohibitively expensive since the memory of the ViT scales with $\mathcal{O}(H^2 W^2)$. Another option could be to upscale the features directly by a factor of $C$ using bilinear-interpolation. This approach computes sub-optimal pixel-level features as it simply bilinearly redistributes the features between the centers of patches.
Such an approach fails to take advantage of the information that is readily available in the original image space. LiFT takes advantage of this information.

\subsection{Lightweight Feature Transform (LiFT)}

\begin{figure*}[!t]
    \centering
    \includegraphics[width=\linewidth,trim={0.7cm 0 0.1cm 0},clip]{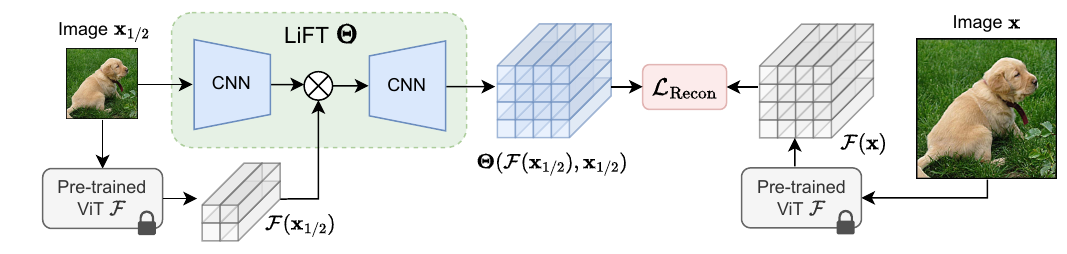}
    \caption{Illustration of \textbf{LiFT}, our proposed \textbf{Li}ghtweight \textbf{F}eature \textbf{T}ransform for generating dense ViT descriptors. The frozen ViT backbone is used to extract features for both low- and high-resolution images. The low-resolution image and its corresponding features are passed through LiFT, which generates a dense version of the features. The LiFT Block first encodes fine-resolution image features using a small CNN. It then combines the CNN features with the ViT features at multiple phases in an upsampling CNN, which outputs dense features. The LiFT block is trained using a self-supervised reconstruction error with the corresponding high-resolution features.}
    \label{fig:arch}
\end{figure*}

Our proposed approach, LiFT, builds on the hypothesis that, even though the ViT feature descriptors have a low spatial resolution, their high dimensionality allows them to store rich information about the image structure. This hypothesis is further supported by works on internal learning of images~\cite{shocher2018zero,glasner2009super,zontak2011internal}. However, unlike internal learning, we propose to train a general-purpose, lightweight upsampling network using only self-supervision to double the resolution of feature descriptors obtained from a frozen, pretrained ViT. Furthermore, we can gain additional fine-level information directly from the original image at the same resolution used to generate the ViT features. We fuse these two information sources to create our final LiFT module, as illustrated in \Cref{fig:arch}.
Given ViT descriptors of size $\frac{H}{P} \times \frac{W}{P}$, a single LiFT expansion block scales them to $\frac{2H}{P} \times \frac{2W}{P}$ in a single forward pass. Our LiFT block is built following a U-Net-style structure \cite{ronneberger2015u} with skip connections, where semantically rich but coarse ViT features are combined with shallow but dense image features derived from a second input with the original image. The image input is processed through a series of convolution blocks and the resulting features are concatenated to the ViT features. Then, a single transpose convolution block is applied to generate the upscaled semantically rich features. Thanks to its fully convolutional nature, the LiFT block can be applied to any image size. One can even apply a LiFT block multiple times to further upscale the features, as shown in Section \ref{sec:prop}.

\subsection{Training Objective}
\label{sec:loss}

Given an image $\x \in \mathbb{R}^{H \times W \times 3}$ and a pretrained, frozen ViT model $\vit$ of stride $P$, we extract last-layer features $\vit(\x) \in \mathbb{R}^{\frac{H}{P} \times \frac{W}{P} \times D}$, where $D$ is the feature dimension. The LiFT block $\lift$ upscales the features from $\frac{H}{P} \times \frac{W}{P}$ to $\frac{2H}{P} \times \frac{2W}{P}$. For training, we propose the following multi-scale reconstruction objective:
\begin{align}
\label{eq:loss}
    \mathcal{L}_\text{Recon} =& \quad d\left( \vit(\x), \lift(\vit(\x_{1/2}), \x_{1/2}\right) + d\left(\vit(\x_{1/2}), \lift(\vit(\x_{1/4}), \x_{1/4}\right)
\end{align}
Where $\x_{1/2}$ and $\x_{1/4}$ are the images resized to $1/2$ and $1/4$ of the original image resolutions, and $d$ is a distance function. For $d$ we select the cosine distance metric, as it inherently normalizes the output and empirically achieves better performance. When processing $\x_{1/2}$ and $\x_{1/4}$, we follow the method of \cite{caron2021emerging} to handle positional embeddings for images of different sizes.
We optimize the parameters of the LiFT module $\lift$ to minimize Equation \ref{eq:loss}.

\subsection{Training Details}
\label{sec:training}

Training LiFT is fast and efficient, as it is lightweight and does not require propagating gradients through the ViT backbone. The LiFT module, $\lift$, is a small network with $1.2M$ trainable parameters. We train LiFT for 5 epochs on the ImageNet dataset on a single GPU. We use a learning rate of 0.001 with a batch size of 256 for stride 16 training.
We use DINO~\cite{caron2021emerging} ViT-S/16 as our base ViT, and we apply color jitter as the only augmentation. In total, training takes only ${\sim}8$ hours on one RTX A6000 GPU. Once our LiFT module is trained, it is a general purpose feature enhancement module that can be directly applied to a range of downstream tasks without need for any further finetuning.
We demonstrate the flexibility of LiFT by applying features from the same DINO+LiFT model to several tasks in Section \ref{sec:perf}.
We present additional analysis of different design choices for training and inference in Appendix A.

\subsection{Using LiFT with Downstream Modules}
\label{sec:vitdet_lift}

Our LiFT module can be easily applied to downstream tasks that directly operate on the output image features.
However, LiFT can also be applied in circumstances where additional downstream modules follow the feature extracting backbone. To demonstrate this, we show how to apply LiFT to the ViTDet architecture~\cite{li2022exploring}, which combines a ViT backbone with a detector head to perform COCO Object Detection and Segmentation. To achieve this, we inject the LiFT module after the backbone and before the head in a pretrained ViTDet model. We train this LiFT module on COCO training data using the same self-supervised objective described in Section \ref{sec:loss}.
We then briefly finetune the pretrained head on the LiFT features.
We present these results in Section \ref{sec:perf}, and we provide additional details of our ViTDet+LiFT architecture in Appendix B.

\subsection{Baseline Methods}
In addition to the base DINO features, we compare to three types of baselines. The first are baselines which finetune DINO features to enhance them without increasing their density. This includes Leopart~\cite{ziegler2022self} and SelfPatch~\cite{Yun_2022_CVPR}. The second are basic feature upsampling methods, which includes bilinear interpolation (BL), resize-convolutions (RC), and Joint Bilinear Upsampling (JBU). For these methods, we double the feature resolution to match LiFT. We use the generalized JBU implementation of \cite{fufeatup}, which includes learnable parameters. We train both JBU and RC following the same optimization protocol as our LiFT module. The final type are dense ViT feature extraction methods, specifically increased image resolution and reduced stride \cite{amir2021deep}.
Note that LiFT represents an orthogonal direction of ViT improvement that can be used in combination with methods like \cite{amir2021deep} and \cite{ziegler2022self}, which we show in the following sections and in Appendix C.
\section{Performance Benefits of LiFT}
\label{sec:perf}

\begin{table*}[t]
\begin{minipage}[t]{\linewidth}
\setlength{\cmidrulewidth}{0.01em}
\renewcommand{\tabcolsep}{6pt}
\renewcommand{\arraystretch}{1.1}
\caption{Comparison between LiFT and other baselines on the Keypoint Correspondence task on SPair-71k. We report PCK@0.1/0.05/0.01 at multiple input resolutions.}
\label{tab:spair_main}
\resizebox{\linewidth}{!}{
\begin{tabular}{@{}lcccccccccccc@{}}
\toprule
& \multicolumn{4}{c}{PCK@0.1} & \multicolumn{4}{c}{PCK@0.05} & \multicolumn{4}{c}{PCK@0.01} \\
\cmidrule[\cmidrulewidth](l){2-5} \cmidrule[\cmidrulewidth](l){6-9} \cmidrule[\cmidrulewidth](l){10-13}
Method/Res. & 56 & 112 & 224 & 448 & 56 & 112 & 224 & 448& 56 & 112 & 224 & 448\\
\midrule
DINO & 2.04	&12.67	&24.76	&28.6	&0.51	&3.61	&9.54	&15.33 & 0.01 &	0.2	&0.54	&1.4 \\
Leopart & 2.35	&11.2	&23.33	&26.54	&0.6	&3.22&	8.9	&12.26 & 0.05 &	0.1&	0.47&	0.79 \\
SelfPatch	& 2.13	&12.18	&23.03&	27.34	&0.44	&3.61&	9.32&	14.44 & 0.02	&0.17	&0.42	&1.12\\
\hdashline
DINO+BL	& 3.81&	13.71	&26.72	&30.48	&1.01&	4.23&	11.37	&16.75	&0.02	&0.14	&0.5	&1.64 \\
DINO+RC	& 3.87	&14.02	&26.09	&30.21&	0.94	&4.41&	11.51	&16.76	&0.03&	0.16	&0.51	&1.7 \\
DINO+JBU & 3.08	&12.26&	24.87&	29.11	&0.82&	3.82&	10.6&	16.01	&0.03&	0.17&	0.47&	1.42 \\
\hdashline
DINO+LiFT & \textbf{5.05} & \textbf{17.72}	&\textbf{28.68}	&\textbf{31.38}&	\textbf{1.19}&	\textbf{6.29}	&\textbf{14.72}&	\textbf{18.90} & \textbf{0.06}&	\textbf{0.29}&	\textbf{0.91}&	\textbf{2.52}\\
\bottomrule
\end{tabular}
}
\end{minipage}
\end{table*}

\myparagraph{SPair Keypoint Correspondence.}
This task involves matching keypoints between pairs of images in the SPair-71k dataset. We follow the evaluation protocol of Amir et al.~\cite{amir2021deep} and report Percentage of Correct Keypoints (PCK) with three different distance thresholds.
We extract dense features using the frozen DINO+LiFT combination trained in Section \ref{sec:training}.
For this task, the features of all methods are resized to match the original image resolution using Lanczos interpolation before feature matching begins.
Table~\ref{tab:spair_main} presents the results.
Compared to both the base DINO model and other baselines, LiFT performs the best across all resolutions for all metrics. For lower resolutions like $56\times56$, LiFT more than doubles the performance of DINO. Also note that under the PCK@0.1 metric, DINO+LiFT at $224\times224$ resolution beats DINO run at $448\times448$, even though both configurations produce final features of the same density.

\myparagraph{DAVIS Video Object Segmentation.}
This task involves propagating a video object segmentation across multiple frames where the first frame ground truth segmentation mask is provided. This is achieved through dense feature matching between frames.
Again, we extract dense features with the same pre-trained DINO+LiFT.
We follow the evaluation protocol of \cite{jabri2020space} and, for brevity, we report results for the J \& F mean metric in the main paper, but we can see consistent improvements across the J mean and F mean individually as shown in Appendix D.
In Table~\ref{tab:davis_main}, it can be seen that across all resolutions LiFT outperforms all other approaches.
At the lowest resolution of $56\times56$, the performance gain over the base DINO is ${\sim}1.75\times$. On average, we improve by $9.4$ points over base DINO.

\begin{table*}[!t]
\begin{minipage}[t]{0.48\linewidth}
\setlength{\cmidrulewidth}{0.01em}
\renewcommand{\tabcolsep}{8.65pt}
\renewcommand{\arraystretch}{1.2}
\caption{DAVIS Video Object Segmentation results for the J \& F Mean metric.}
\label{tab:davis_main}
\resizebox{\linewidth}{!}{
\begin{tabular}{@{}lcccc@{}}
\toprule
Method/Res. & 56 & 112 & 224 & 448 \\
\midrule
DINO & 7.4	&17.5	&33.0	&50.9	\\
Leopart	& 6.9	&16.1	&30.3	&45.1 \\
SelfPatch & 7.4	&17.2&	33.0	&51.4	\\
\hdashline
DINO+BL &10.8	&23.7&	37.0	&53.0\\
DINO+RC & 11.0	&24.0	&37.4	&53.2 \\
DINO+JBU & 11.2	&26.2&	39.0&	54.5 \\
\hdashline
DINO+LiFT & \textbf{13.0}&	\textbf{28.0}	& \textbf{44.3}	&\textbf{61.1} 	\\
\bottomrule
\end{tabular}
}
\end{minipage}
\hfill
\begin{minipage}[t]{0.48\linewidth}
\setlength{\cmidrulewidth}{0.01em}
\renewcommand{\tabcolsep}{6pt}
\renewcommand{\arraystretch}{1.2}
\caption{COCO20K Unsupervised Object Discovery results for CorLoc metric.}
\label{tab:res_disc}
\resizebox{\linewidth}{!}{
\begin{tabular}{@{}lcccc@{}}
\toprule
Method/Res. & 56 & 112 & 224 & 448 \\
\midrule
DINO	&16.28	&40.08&	53.98&	57.99\\
Leopart	&16.14&	26.78	&43.89	&44.08\\
SelfPatch	&14.15&	35.76&	52.18&	55.47\\
\hdashline
DINO+BL	& 17.78 & 35.62 & 51.53 & 56.84 \\
DINO+RC	& 22.92 & 42.53 & 54.52 & 58.40 \\
DINO+JBU & 21.36 & 43.87 & 55.45 & 58.82 \\
\hdashline
DINO+LiFT	&\textbf{27.72}&	\textbf{50.20}&	\textbf{58.03}&	\textbf{60.50}\\
\bottomrule
\end{tabular}
}
\end{minipage}
\end{table*}
\begin{table}[t]
\begin{minipage}[t]{\linewidth}
\setlength{\cmidrulewidth}{0.01em}
\renewcommand{\tabcolsep}{5pt}
\renewcommand{\arraystretch}{1.2}
\caption{COCO Detection and Segmentation results with ViTDet+LiFT for two different heads: Mask R-CNN (MR) and Cascade R-CNN (CR).}
\label{tab:res_coco}
\centering
\resizebox{0.7\linewidth}{!}{
\begin{tabular}{@{}lcccccc@{}}
\toprule
& \multicolumn{3}{c}{Detection} & \multicolumn{3}{c}{Segmentation} \\

\cmidrule[\cmidrulewidth](l){2-4} \cmidrule[\cmidrulewidth](l){5-7}
Method & AP &  AP$_{50}$ & AP$_{75}$ & AP &  AP$_{50}$ & AP$_{75}$   \\
\midrule
ViTDet(MR)		& 39.50 & 61.56 & 42.23 & 37.81 & 60.47 & 39.97 \\
ViTDet(MR)+LiFT	& \textbf{45.98} & \textbf{66.41} & \textbf{49.87} 	& \textbf{40.78} & \textbf{63.34} & \textbf{43.57} \\
\hdashline
ViTDet(CR)		& 45.65 & 64.39 & 48.90 & 40.93 & 62.86 & 44.28 \\
ViTDet(CR)+LiFT	& \textbf{47.92} & \textbf{65.88} & \textbf{51.67} 	& \textbf{41.42} & \textbf{63.2}1 & \textbf{44.45} \\
\bottomrule
\end{tabular}
}
\end{minipage}
\end{table}

\myparagraph{Unsupervised Object Discovery.}
We test the benefits of LiFT for Unsupervised Single Object Discovery on COCO20K~\cite{Lin2014MicrosoftCC}. 
We apply TokenCut~\cite{wang2023tokencut}, which performs Graph Cut on features, to LiFT and the other baseline methods. 
Similar to prior works~\cite{wang2023tokencut, cho2015unsupervised,deselaers2010localizing,wei2019unsupervised,rambhatla2021pursuit, vo2021large} we report the  Correct Localization (CorLoc) metric, which is computed as the fraction of the images in which at least one object box prediction has an IoU greater than a threshold (0.5) with a ground truth box.
As shown in Table \ref{tab:res_disc}, LiFT gives a good boost in CorLoc, with gains across all resolutions and with LiFT outperforming the other approaches. It should be noted that the performance gains at lower resolutions are especially large. For example, at the $56\times56$ resolution, we see an improvement of $11.44$ over DINO.
In Appendix D, we present additional results on PASCAL VOC 2007~\cite{pascal-voc-2007} and PASCAL VOC 2012~\cite{pascal-voc-2012}, and we observe similar overall trends.

\myparagraph{COCO Detection and Segmentation.}
To further demonstrate the versatility of LiFT, we show how it can be applied to COCO Detection and Segmentation using a ViTDet+LiFT model as described in Section \ref{sec:vitdet_lift}.
We present results for two ViTDet variants using different heads: Mask R-CNN (MR)~\cite{he2017mask}, and Cascade R-CNN (CR)~\cite{cai2018cascade}.
In Table \ref{tab:res_coco}, ViTDet(MR)+LiFT gives a ${\sim}6.5\%$ boost for Detection AP and a ${\sim}3\%$ boost for Segmentation AP.
Smaller but still consistent boosts are also achieved when combining LiFT with ViTDet(CR).
\section{Computational Efficiency of LiFT}
\label{sec:eff}

We have shown that feature densification with LiFT provides significant performance benefits in several tasks. We now show why LiFT is a \textit{Lightweight} transform, as it is vastly more computationally efficient than other alternatives that increase the token density of the ViT backbone. 
A trivially easy way to boost the density of ViT features is to increase the resolution of the input image.
Another option is the dense feature extraction strategy of \cite{amir2021deep} which increases the number of tokens in the network by reducing the stride during patch extraction.
Both of these methods increase token density but at a high computational cost.
We present a comprehensive compute cost \vs performance benefit analysis for LiFT against and in combination with these methods, and we show that LiFT acts as a ``shortcut'' to achieve higher resolution features for minimal extra compute.

\begin{table*}[!t]
\begin{minipage}[t]{0.51\linewidth}
\setlength{\cmidrulewidth}{0.01em}
\renewcommand{\tabcolsep}{2pt}
\renewcommand{\arraystretch}{1.3}
\caption{FLOPs comparison for a single forward pass across resolutions and strides for DINO and DINO+LiFT. Along with FLOPS, we report the performance on both SPair-71k and DAVIS.}
\label{tab:computation}
\resizebox{\linewidth}{!}{
\begin{tabular}{@{}lccccc@{}}
\toprule
Method & Res. & Stride & FLOPs (G) & PCK@0.1 & J\&F mean\\

\midrule
DINO & \multirow{4}{*}{224}	& \multirow{2}{*}{16}	& 4.34	& 24.76 & 33.0\\
DINO+LiFT &	&  & 5.30 & 28.68 & 44.3\\
DINO & & \multirow{2}{*}{8} & 16.07	& 29.92 &	43.9\\
DINO+LiFT &  &  & 19.65	&31.91	& 52.6 \\ \hdashline
DINO & \multirow{4}{*}{448}	& \multirow{2}{*}{16}	&17.28	& 28.60 & 50.9\\
DINO+LiFT &  &  & 21.12	&31.38 & 61.1\\
DINO & &	\multirow{2}{*}{8} &66.60	&31.92	& 61.9\\
DINO+LiFT &  &  & 81.18	&32.20& 69.7\\
\bottomrule
\end{tabular}
}
\end{minipage}
\hfill
\begin{minipage}[t]{0.45\linewidth}
\centering
    \captionof{figure}{Performance \vs Compute Cost trade-off curve for SPair-71k keypoint correspondence. For any given FLOP-budget, DINO+LiFT achieves far superior performance.}
    \centering
    \includegraphics[width=\linewidth]{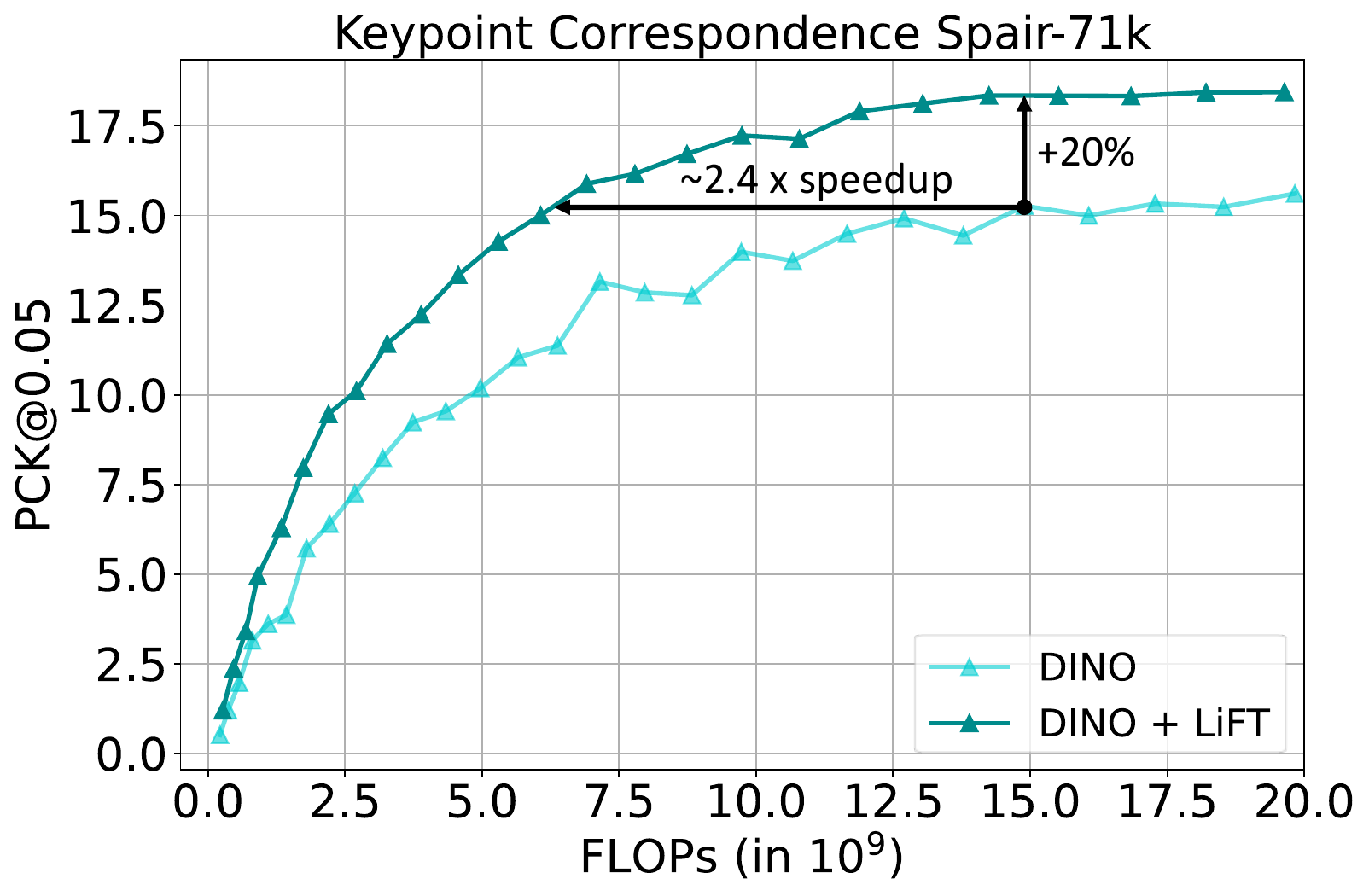}
    \label{fig:tradeoff}
\end{minipage}
\end{table*}

\myparagraph{LiFT, Resolution, and Stride.}
For this analysis, we take SPair Keypoint Correspondence and DAVIS Video Segmentation as the tasks, and we measure compute cost in FLOPs (G) against performance both with and without LiFT over a range of input resolutions and strides.
As shown in rows 1 and 2 of Table~\ref{tab:computation}, LiFT introduces a small increase in FLOPs from $4.34$G to $5.30$G ($+22\%$) but also brings a significant performance improvement of $3.92$ and $11.3$ points on the PCK@0.1 and J \& F Mean metrics respectively. For comparison, reducing the stride from $16$ to $8$ (row 1 \vs row 3) gives a similar level of improvement, but nearly quadruples the FLOPs from $4.34$G to $16.07$G ($+270\%$).
A similar trend of large improvements can be seen when comparing pairs of rows (3 \vs 4, 5 \vs 6) with DINO \vs DINO+LiFT. The best overall results are achieved by combining LiFT with increased resolution and reduced stride, but this also comes with the highest computation cost. It should also be noted that DINO+LiFT at $224\times224$ resolution and stride 8 (row 4) uses $3\times$ less FLOPs than DINO at $448\times448$ resolution at stride 8 (row 7) while having similar PCK performance.

\myparagraph{Performance Trade-Off Curve.}
We present a comprehensive Compute Cost \vs Performance trade-off curve for SPair Keypoint Correspondence in Figure \ref{fig:tradeoff}. By incrementally increasing the input resolution, we can gradually increase both the performance and inference cost for DINO and DINO+LiFT features.
We find that LiFT significantly outperforms the base DINO features at any given FLOP allowance, in most cases seeing a ${\sim}20\%$ performance gain.
Alternatively, LiFT can be run at a lower input resolution to achieve equivalent performance at a fraction of the compute cost. For example, to achieve a score of $15.0$ in PCK@0.05, DINO+LiFT only requires ${\sim}6.25$ Giga-FLOPs of compute power, while the standard DINO requires at least $15$ Giga-FLOPs.

\begin{table*}[!t]

\begin{minipage}[t]{0.59\linewidth}
\setlength{\cmidrulewidth}{0.01em}
\renewcommand{\tabcolsep}{2pt}
\renewcommand{\arraystretch}{1.5}
\caption{Comparison of parameters \& FLOPs \vs performance at $224\times224$ resolution for Keypoint Correspondence. We report PCK@0.1 and PCK@0.05 on SPair-71k.}
\label{tab:reb3}
\centering
\resizebox{\linewidth}{!}{
\begin{tabular}{@{}lccccc@{}}
\toprule
Method & Parameters & FLOPs (G) & PCK@0.1 & PCK@0.05\\
\midrule
DINO S/16	    & 21M & 4.34 &   24.76 & 9.54 \\
DINO S/16+LiFT	& 22.2M & 5.30 & \textbf{28.68}  & \textbf{14.72}\\
DINO B/16	& 85M & 17.21 & 24.90 & 9.64\\
\bottomrule
\end{tabular}
}
\end{minipage}
\hfill
\begin{minipage}[t]{0.37\linewidth}
\setlength{\cmidrulewidth}{0.01em}
\renewcommand{\tabcolsep}{4pt}
\renewcommand{\arraystretch}{1}
\caption{LiFT when using different backbones for training, inference, or both. We report PCK@0.1 on SPair-71k.}
\label{tab:cross_model}
\centering
\resizebox{\linewidth}{!}{
\begin{tabular}{@{}lcccc@{}}
\toprule
& & \multicolumn{3}{c}{Training Model} \\
\cmidrule[\cmidrulewidth](l){3-5}
\makecell{Inference\\Model} & No LiFT & DINO & MoCo & ViT \\
\midrule
DINO & 28.6	& \textbf{31.38}	&16.02 & 20.71	\\
MoCo	& 12.31	& 9.86	&\textbf{16.34} & 11.31 \\
ViT & 16.9	& 12.55	& 8.91	& \textbf{18.69}	\\
\bottomrule
\end{tabular}
}
\end{minipage}
\end{table*}

\myparagraph{Parameter Count.}
We acknowlege that the addition of the LiFT module slightly increases the overall model size and parameter count as shown in Table \ref{tab:reb3}. However, this addition is quite small and only represents a $+5.7\%$ change in total parameters. For comparison, the jump from ViT-S to ViT-B results in a $+304\%$ increase in parameters.
Furthermore, for dense tasks like SPair Keypoint Correspondence, we find that the performance benefits provided by LiFT far exceed the benefits of a larger backbone. 
Note that the methods Leopart and SelfPatch do not increase the parameter count of the ViT, as they finetune the DINO backbone instead of introducing new modules.
We believe the major performance benefits of LiFT justify the small extra costs in Parameter Count and Inference FLOPs for a given resolution.
And, as shown in the previous section, for any fixed FLOP-budget, DINO+LiFT achieves far superior performance.
\section{Properties of LiFT}
\label{sec:prop}

\begin{figure*}[t]
    \centering
    \begin{subfigure}[b]{0.261111111\textwidth}
        \centering
        \includegraphics[width=\textwidth]{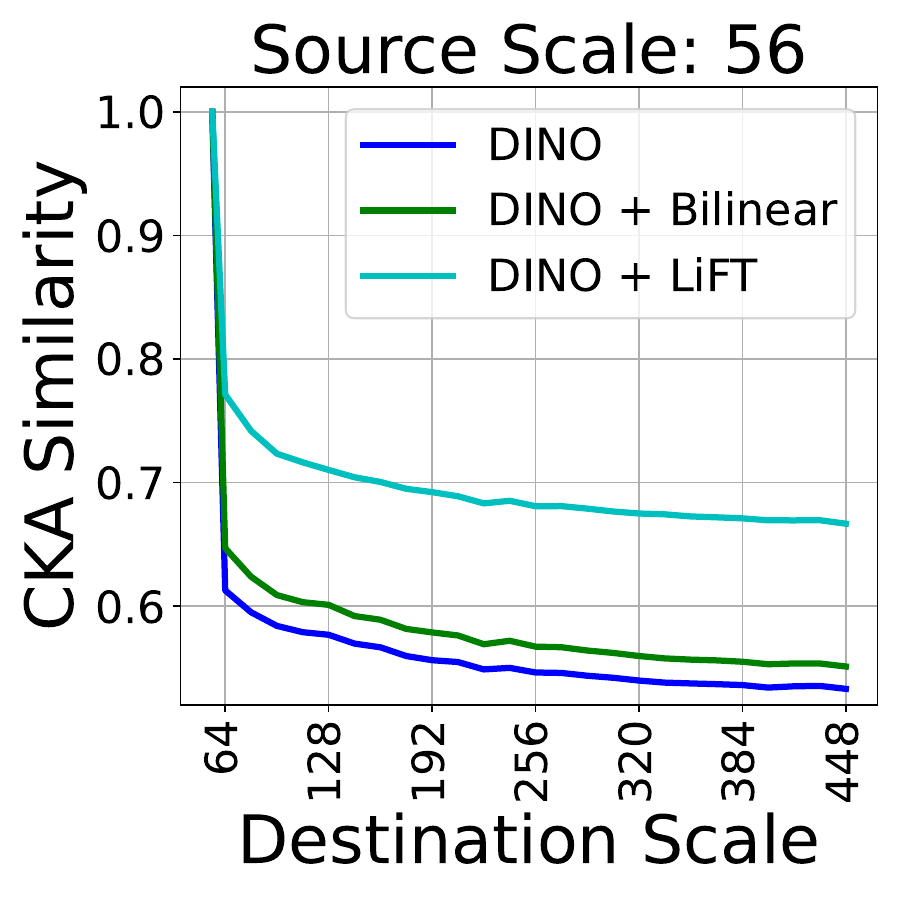}
        \label{fig:cka-lines-56}
    \end{subfigure}
    \hfill
    \begin{subfigure}[b]{0.235\textwidth}
        \centering
        \includegraphics[width=\textwidth]{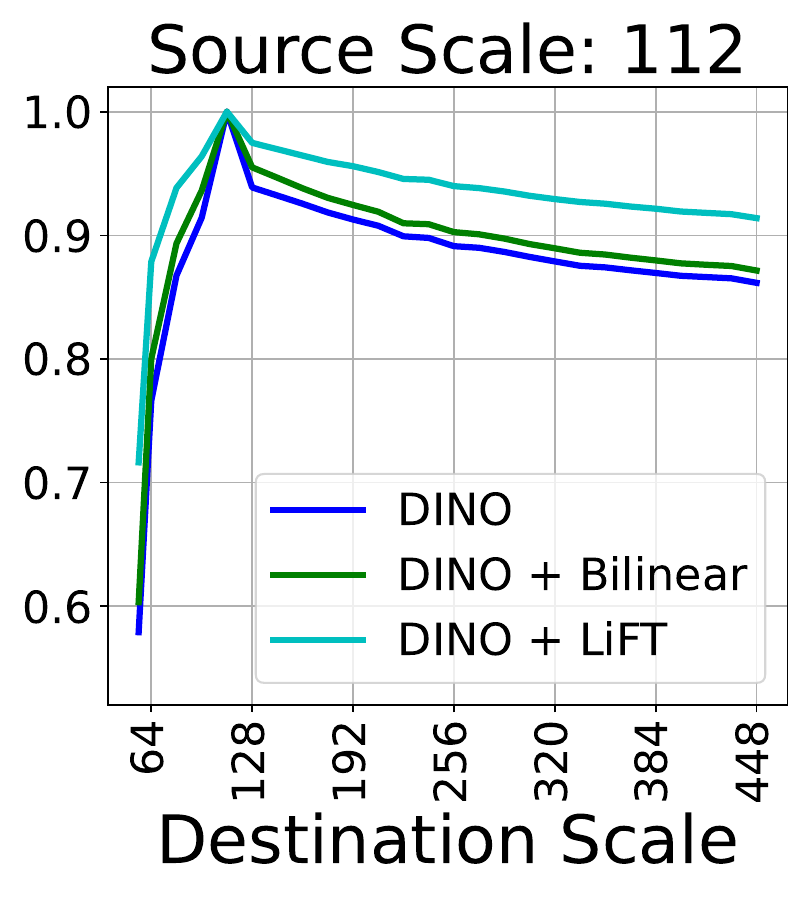}
        \label{fig:cka-lines-112}
    \end{subfigure}
    \hfill
    \begin{subfigure}[b]{0.235\textwidth}
        \centering
        \includegraphics[width=\textwidth]{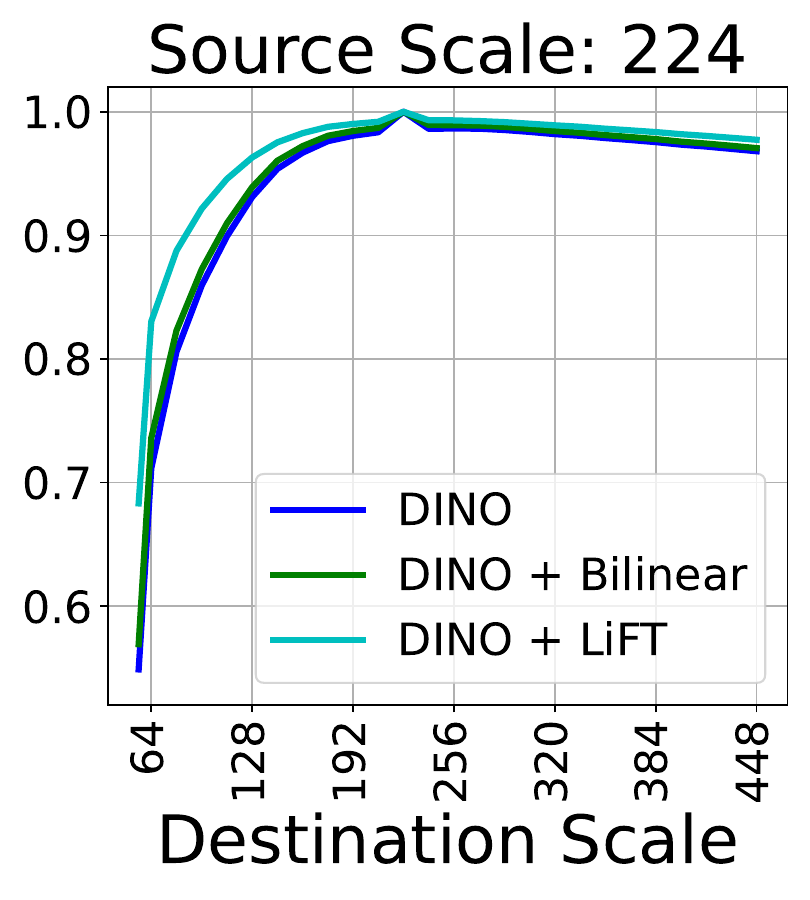}
        \label{fig:cka-lines-224}
    \end{subfigure}
    \hfill
    \begin{subfigure}[b]{0.235\textwidth}
        \centering
        \includegraphics[width=\textwidth]{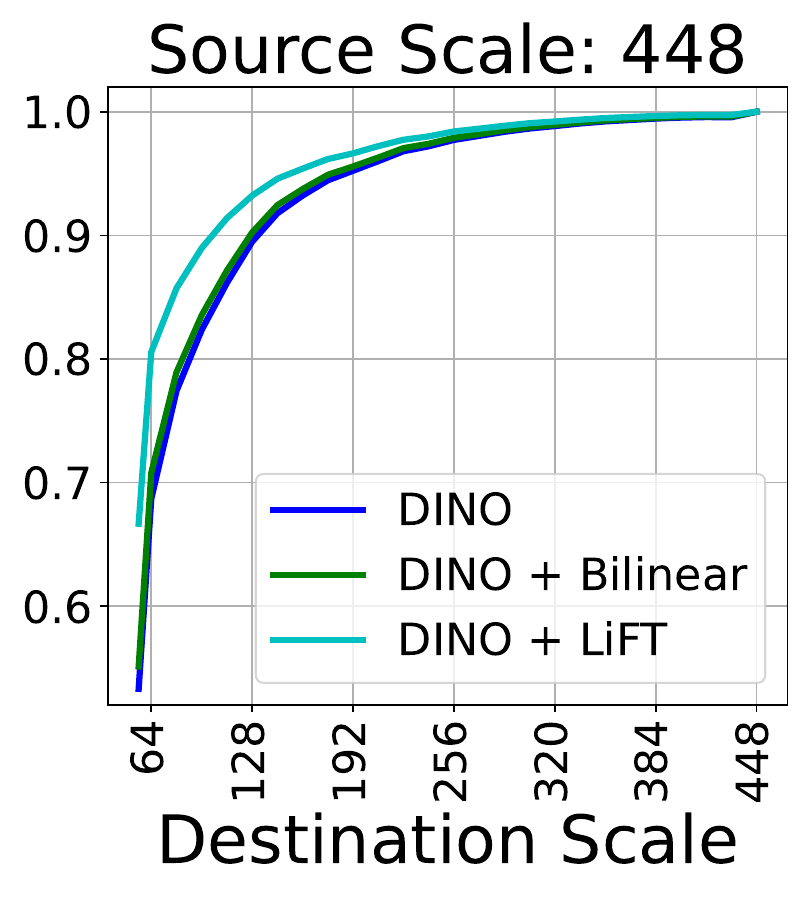}
        \label{fig:cka-lines-448}
    \end{subfigure}
    \caption{CKA Similarity of ViT features extracted from SPair-71k images at different input image sizes, denoted by Source Scale and Destination Scale. LiFT produces features that are more scale-invariant, especially for smaller scale inputs and objects.}
    \label{fig:cka-lines}
\end{figure*}

\myparagraph{LiFT and Scale Invariance of Features.}
In this experiment, we demonstrate that LiFT intrinsically learns to generate features that are more scale-invariant. We use the Centered Kernel Alignment (CKA) metric~\cite{cortes2012algorithms, kornblith2019similarity, subramanian2021torch_cka}, which can measure the similarity of a pair of feature maps even when they are different sizes. Using images in SPair-71k, we re-scale each image to a range of different sizes and then extract features with DINO or DINO+LiFT and measure the CKA similarity between all input size pairings.
As a baseline, we also compare with bilinearly upsampled DINO features.
In Figure \ref{fig:cka-lines}, we take four source scales and plot the CKA similarity with the features at all other scales.
We see that LiFT greatly improves the inter-scale feature similarity for small input scales, and moderately improves the similarity for medium and larger scales. This shows that LiFT produces representations that are more scale-invariant than the base DINO features.
Bilinear upsampling does provide a small improvement in CKA similarity across scales, though the benefit is far smaller.
This scale invariance property of LiFT is learned automatically, and likely comes from its multi-scale reconstruction objective.
LiFT must learn to counteract the effect of input scale to map features from low resolution inputs to those of high resolution inputs. This property is desirable for dense tasks where objects appear at different scales.

\begin{figure}[t]
  \centering
    \includegraphics[width=\linewidth]{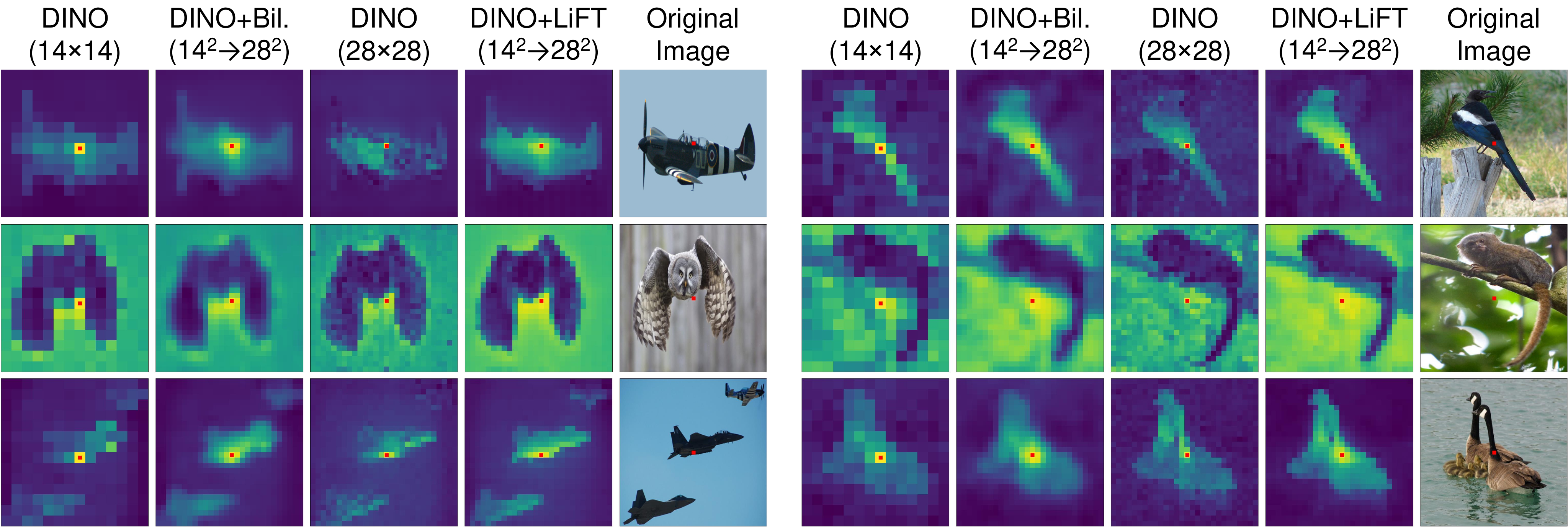}
    \caption{Visualization of the self-similarity of features for DINO, DINO+Bilinear interpolation, DINO with higher resolution image, and DINO+LiFT. To generate this visualization, the self-similarity is computed using the feature corresponding to the center of the grid (marked in red) and all other features from each spatial location. Brighter map shows a higher similarity. Best viewed digitally in color.}
    \label{fig:qualitative}
\end{figure}

\myparagraph{Enhanced Self-Similarity Maps with LiFT.}
To further improve our understanding of the dense features generated by LiFT, we visualize the self-similarity of the features in Figure~\ref{fig:qualitative}. Base DINO S/16 yields $14\times14$ features for a $224\times224$ image. In our comparisons, we visualize these features alongside: a bilinearly interpolated upsampling of these same features to $28\times28$ (DINO+Bilinear), DINO features for a $448\times448$ input image, and finally DINO+LiFT features generated for a $224\times224$ input image. The last three configurations all yield $28\times28$ feature grids. To visualize these features, we select the center-most token feature for each of the maps and compute its similarity with all other features and visualize the similarity scores.
We show six feature similarity maps in Figure~\ref{fig:qualitative}, and additional sample visualizations can be found in Appendix E.
We find that, qualitatively, the LiFT output gives a cleaner boundary and better highlights the content corresponding to the central patch.
Having a high similarity to relevant regions and a clear boundary is beneficial for multiple localized downstream tasks. It also indicates that the DINO+LiFT features have better spatial awareness of object boundaries. Note that for row 2, where the central pixel corresponds to the background region, the similarity map highlights the background in the image, though it still appears that DINO+LiFT produces better similarity maps with sharper edges. In row 3, when there are multiple objects of the same type, DINO+LiFT better highlights the separate object instances.
These results are qualitative, but they suggest that the LiFT-enhanced features have better content and boundary information than the base DINO features, which likely contributes to their improved performance in correspondence and segmentation tasks.

\myparagraph{Variations in Backbone.}
One of the requirements of a feature densifying approach is that it should be easy to train on any backbone.
To this end, we show that our approach consistently gives a performance gain with multiple different backbones: DINO (Table~\ref{tab:ablation} row 1 \vs row 2), MoCo (Table~\ref{tab:ablation} row 4 \vs row 5) and a fully-supervised ViT (Table~\ref{tab:ablation} row 7 \vs row 8). This consistent improvement shows that LiFT can be trained in the exact same manner on differently trained ViTs without need for careful hyperparameter tuning. To verify that LiFT does not simply learn a bilinear interpolation, we apply LiFT modules trained on one backbone to the output of a different backbone. We show these results in Table~\ref{tab:cross_model} using the PCK@0.1 metric and $448\times448$ images.
It can be seen that when LiFT is applied to a different model than what was used to train it, the performance drops and is lower than not applying LiFT. This shows that LiFT learns a model-specific feature-densifying transform and not a simple interpolation. 

\begin{table*}[!t]
\begin{minipage}[t]{0.57\linewidth}
\setlength{\cmidrulewidth}{0.01em}
\renewcommand{\tabcolsep}{8pt}
\renewcommand{\arraystretch}{1}
\caption{Performance for LiFT with different backbones and for repeated application of LiFT. We report PCK@0.1 on SPair-71k.}
\label{tab:ablation}
\centering
\resizebox{1.0\linewidth}{!}{
\begin{tabular}{@{}cllcccc@{}}
\toprule
\multirow{2}{*}{Row}&\multirow{2}{*}{Backbone} & \multirow{2}{*}{Method}& \multicolumn{4}{c}{Resolution}\\
\cmidrule[\cmidrulewidth](l){4-7} 
 &  &  & 56 & 112 & 224 & 448\\

\midrule
1 & \multirow{3}{*}{DINO} & - & 2.04 &	12.67&	24.76&	28.6\\
2 & & LiFT & 5.05 &	17.72 &	28.68 &	\textbf{31.38}\\
3 & & 2$\times$LiFT & \textbf{7.42} &\textbf{20.12}&	\textbf{29.45}	& 31.35 \\ 
\hdashline							
4 &\multirow{3}{*}{MOCO}& - &1.27&	3.43&	7.37&	12.31\\
5 & &LiFT	& 6.48&	10.51&	14.13&	16.34\\ 
6 & &2$\times$LiFT 	&\textbf{8.72}&	\textbf{13.21}	&\textbf{16.12}	&\textbf{17.08}\\ 
\hdashline		
7 & \multirow{3}{*}{ViT}& - & 1.26 &	5.72	&13.23	&16.9\\
8 & &LiFT	& 3.76&	9.21&	\textbf{16.5}8&	\textbf{18.69}\\
9 & & 2$\times$LiFT &\textbf{5.17}&	\textbf{9.89}	& 16.49	& 18.18\\
\bottomrule
\end{tabular}
}
\end{minipage}
\hfill
\begin{minipage}[t]{0.39\linewidth}
\centering
    \captionof{figure}{With no extra training, repeated application of LiFT produces pixel-dense feature maps.}
    \centering
    \includegraphics[width=\linewidth]{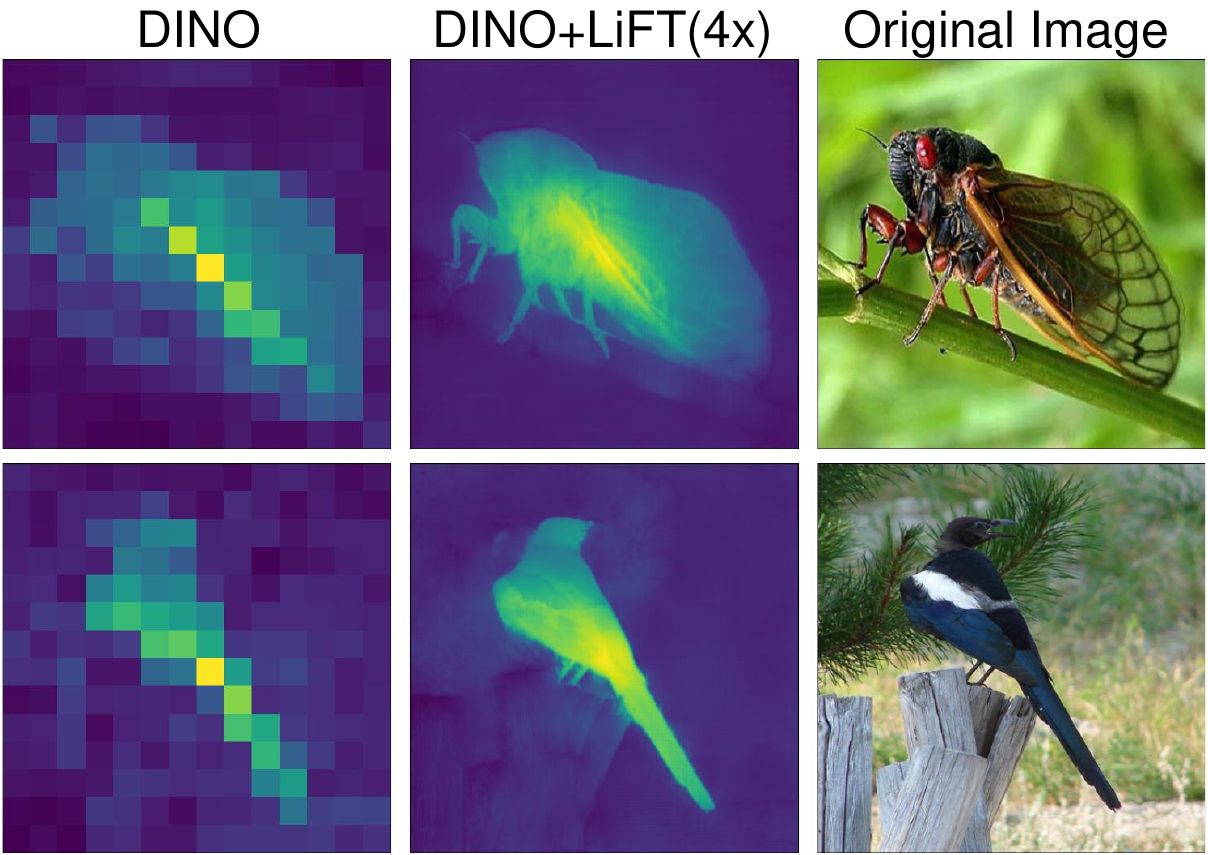}
    \label{fig:pixelscale}
\end{minipage}
\end{table*}

\myparagraph{Repeated Application of the LiFT Module.}
In our base approach, the LiFT module is applied once to the ViT features to double their resolution. In this section, we also test the potential benefits of applying the LiFT block multiple times.
To check this, we take the LiFT module and super-resolute the features twice by passing the output of first super-resolution again through the same LiFT network. We denote this approach as `2$\times$LiFT' in Table~\ref{tab:ablation}.
We can see that in most cases the performance increases after recursively applying LiFT. This is especially true for the $56\times56$, $112\times112$, and $224\times224$ resolutions. For $448\times448$, there are a few places where it negligibly drops performance, but it still shows improvement for most cases.
Repeated application of LiFT can also be used to create detailed, pixel-dense feature self-similarity maps, as shown in Figure \ref{fig:pixelscale} for a $4\times$ repeated application of the same pretrained LiFT module. Such maps could be useful for tasks like Unsupervised  Semantic Segmentation \cite{gao2022luss}.
\section{Conclusion and Discussion}
We have presented \textbf{LiFT}, a simple yet effective self-supervised \textbf{Li}ghtweight \textbf{F}eature \textbf{T}ransform to boost the density of features of pretrained ViT backbones.
This approach allows us to extract higher resolution spatial features from ViTs which can then be used for multiple dense downstream tasks.
LiFT is task-agnostic and gives significant boosts in SPair Keypoint Correspondence, DAVIS Video Object Segmentation, Unsupervised Object Discovery, and COCO Detection and Segmentation.
This benefit comes for a fraction of the computational cost compared with other densification methods.
LiFT is a lightweight module that is easily trained with a self-supervised objective, and it is far cheaper than full backbone finetuning, as is done by other prior works.
Through extensive experiments, we have shown that LiFT can be trained easily on any backbone and consistently leads to improved performance by generating better quality, higher-density features. We also show that LiFT can be applied to its own output in a recursive manner enabling good performance with even lower image resolutions.
Finally, we show that our surprisingly simple method has several desirable emergent properties including scale-invariant features, and better object boundary maps. This makes LiFT a useful multipurpose tool for many potential downstream applications.

\myparagraph{Acknowledgements.}
This work was partially supported by NSF CAREER Award (\#2238769) to AS, and the NSF and NIST Institute for Trustworthy AI in Law and Society (TRAILS) (\#2229885). The U.S. Government is authorized to reproduce and distribute reprints for Governmental purposes notwithstanding any copyright annotation thereon. The authors acknowledge UMD’s supercomputing resources made available for conducting this research. The views and conclusions contained herein are those of the authors and should not be interpreted as necessarily representing the official policies or endorsements, either expressed or implied, of NSF, NIST, or the U.S. Government.
We would also like to thank our colleagues Matthew Gwilliam and Pravin Nagar for their feedback on this work.

\bibliographystyle{splncs04}
\bibliography{main}

\newpage
\appendix
\section{Ablation Study of LiFT Design Choices}
\label{sec:apdx_ablation}

\begin{table*}[t]
\begin{minipage}[t]{\linewidth}
\setlength{\cmidrulewidth}{0.01em}
\renewcommand{\tabcolsep}{10pt}
\renewcommand{\arraystretch}{1}
\caption{Ablation of different design decisions for LiFT training for three different ViT backbones. We report PCK@0.1 and PCK@0.05 on SPair-71k. For each backbone, we mark the best score for each metric and input resolution in \textbf{bold}.}
\label{tab:ablationsupp}
\centering
\resizebox{1.0\linewidth}{!}{
\begin{tabular}{@{}llcccccccc@{}}
\toprule
& & \multicolumn{4}{c}{PCK@0.1} & \multicolumn{4}{c}{PCK@0.05} \\
\cmidrule[\cmidrulewidth](l){3-6} \cmidrule[\cmidrulewidth](l){7-10}
Row & Method/Resolution & 56 & 112 & 224 & 448 & 56 & 112 & 224 & 448\\

\midrule
1 & DINO	& 2.04 &	12.67&	24.76&	28.6	&0.51	&3.61	&9.54	&15.33\\
2 &DINO + Random LiFT & 1.45 &	2.37&	4.21&	6.16	&0.35	&0.7	&1.41	&2.35\\
3 &DINO + LiFT No Img.	&4.38&	15.74&	28.49&	\textbf{31.42}&	1.14&	5.03&	13.28&	18.33\\
4 &DINO + LiFT L1	&4.48&	16.64&	27.77	&31.03	&1.01	&5.93&	13.88	&18.09\\
5 &DINO + LiFT L2	&4.82	& \textbf{17.72}	&28.17	&31.13	&\textbf{1.29}	&6.18	&14.12	&18.37\\
6 &DINO  + LiFT & \textbf{5.05} &	\textbf{17.72} &	\textbf{28.68} &	31.38&	1.19&	\textbf{6.29}&	\textbf{14.72} &	\textbf{18.90}\\
\hdashline
								
7 &MOCO & 1.27&	3.43&	7.37&	12.31&	0.21&	0.84&	2.35	&5.49\\
8 &MOCO + Random LiFT & 2.59	&3.08&	4.05	&5.79&	0.67	&0.77	&1.31	&2.1\\
9 &MOCO + LiFT  No Img.	&4.58&	8.78&	13.01&	15.48	&1.18&2.69	&4.95&	7.27\\
10 &MOCO + LiFT L1&	6.12&	9.80	&13.73	&14.98	&1.59&	3.22&	5.86	&7.53\\
11 &MOCO + LiFT L2	&6.37	&10.08	&13.91&\textbf{16.41}	&1.53	&3.12&5.99&	\textbf{8.34}\\
12 &MOCO + LiFT	& \textbf{6.48}&	\textbf{10.51}&	\textbf{14.13}&	16.34&	\textbf{1.74}&	\textbf{3.36}&	\textbf{6.42}&	8.05\\ 
\hdashline
								
13 &ViT 	& 1.26 &	5.72	&13.23	&16.9	&0.27	&1.62	&4.89	&7.34\\
14 &ViT + Random LiFT & 2.36	&3.29&	7.15&	8.21	&0.58	&1.09&	2.33&	3.13\\
15 &ViT + LiFT  No Img.	&2.94&	7.76&	15.69&	18.74&	0.79&	2.22&	5.68&	8.23\\
16 &ViT + LiFT L1	&3.27	&8.32&	16.04&	18.29	&0.79	&2.74	&6.77&	8.45\\
17 &ViT + LiFT L2	&3.57	&8.78	&16.29	&\textbf{18.80}	&0.97	&2.64&	\textbf{6.82}&	\textbf{8.87}\\
18 &ViT + LiFT	& \textbf{3.76}&	\textbf{9.21}&	\textbf{16.5}8&	18.69& \textbf{1.02}&	\textbf{2.71}&	6.63&	8.81\\
\bottomrule
\end{tabular}
}
\end{minipage}

\end{table*}

We present a careful analysis of LiFT design configurations by varying different factors in both training and inference. To show the general applicability and benefits of LiFT, we include three different backbones in this study: DINO \cite{caron2021emerging}, MoCo v3 (MoCo for short) \cite{he2020momentum}, and a Fully Supervised ViT (ViT for short). We standardize the architecture to a ViT S/16 backbone for this analysis.
We use the SPair-71k dataset and the keypoint correspondence task as the main representative metric for this analysis. The results are summarized in Table~\ref{tab:ablationsupp}.

\subsection{Random LiFT} 
One might question if LiFT actually benefits from training, or if the simple act of increasing the feature resolution with any arbitrary function is sufficient to improve performance.
To test this question, we take a random initialization of the LiFT model and measure its performance. We denote this model as `Random LiFT' in Table~\ref{tab:ablationsupp}.
It can clearly be seen in Table~\ref{tab:ablationsupp} rows 2, 8, and 14 that a randomly initialized LiFT model does not do anything meaningful as it performs poorly on all metrics. 
These results validate the importance of LiFT's self-supervised training method.

\subsection{Ablation of Image Input to LiFT}
In our approach, we increase the feature resolution through LiFT by also using the image as a source of finer spatial information.
It should be noted that we use the image at the same resolution as was used to generate the initial features,
which means LiFT does not have or require any additional information beyond the original ViT's input.
To show the importance of this image information, we present a version of LiFT with the image input ablated, denoted as `LiFT No Img.' in \Cref{tab:ablationsupp}.
We can see from rows 3 \vs 6, 9 \vs 12, and 15 \vs 18, that providing the image input helps LiFT produce better quality features which give improved performance on the keypoint correspondence task.
It appears that ablating the image input is less harmful for higher-resolution inputs like 448, which makes intuitive sense as the feature map resolution is higher and thus more detail about the object boundaries can be represented. For DINO and the supervised ViT (rows 3 and 15), the no-image LiFT actually does very slightly better at 448 input resolution for PCK@0.1, but for all other cases normal LiFT is better. For PCK@0.05, the standard LiFT with image input is consistently much better.
We believe this happens because LiFT can take direct cues regarding scene and object boundaries from the image input and generate higher resolution features which better respect these contours.

\subsection{Effect of Distance Function}
In our final approach, we use cosine distance to compute the loss between the ViT-generated higher resolution features and the upscaled features from LiFT.
In \Cref{tab:ablationsupp}, we compare with two alternative options for this distance function, specifically the L1 and L2 distance metrics.
We denote these as `LiFT L1' and `LiFT L2' respectively.
Cosine distance gives the best performance in most cases, such as in rows 4 \& 5 \vs row 6, rows 10 \& 11 \vs row 12, and rows 16 \& 17 \vs row 18. 
For higher-resolution inputs, L2 distance is sometimes slightly better than cosine distance, but in most cases cosine is preferable.
We believe this occurs because of the inherent normalization that cosine distance provides before computing the final loss.

\begin{table*}[b]
    \renewcommand{\tabcolsep}{7.2pt}
    \caption{Ablation of LiFT training epochs on ImageNet, including longer training. Results are shown for DINO+LiFT on Keypoint Correspondence using PCK@0.1.}
    \centering
      \resizebox{0.68\linewidth}{!}{
      \begin{tabular}{@{}lccccc@{}}
        \toprule
        Res/Epochs & $5$ & $10$ & $30$ & $50$ & $100$\\
        \midrule
        $112\times112$ & 17.47& 17.53 &17.75 &17.97 &18.14 \\
        $224\times224$ & 28.45& 28.50& 28.65 &29.00 &29.11 \\
        \bottomrule
      \end{tabular}
      }
    \label{tab:epoch}
\end{table*}

\subsection{Ablation of Training Epochs}
As an additional experiment, we train the LiFT module on ImageNet for an extended period up to 100 epochs on 4 GPUs in Table \ref{tab:epoch}. We find that there are small performance gains from training to very long epochs, however performance mostly saturates by epoch 5. At resolution 224, DINO+LiFT at 5 epochs gives a ${\sim}3.7$ point gain over the base DINO model, while training 95 epochs further only gives an additional $0.66$ point gain. We believe this early saturation is thanks to the LiFT network's small size.

\begin{table*}[t]
\begin{minipage}[t]{\linewidth}
\setlength{\cmidrulewidth}{0.01em}
\renewcommand{\tabcolsep}{7pt}
\renewcommand{\arraystretch}{1.1}
\caption{Application of LiFT to various backbones for the Keypoint Correspondence task on SPair-71k for all metrics. LiFT gives consistent performance improvements.}
\label{tab:backbones_supp}
\resizebox{\linewidth}{!}{
\begin{tabular}{@{}lcccccccccccc@{}}
\toprule
& \multicolumn{4}{c}{PCK@0.1} & \multicolumn{4}{c}{PCK@0.05} & \multicolumn{4}{c}{PCK@0.01} \\
\cmidrule[\cmidrulewidth](l){2-5} \cmidrule[\cmidrulewidth](l){6-9} \cmidrule[\cmidrulewidth](l){10-13}
Method/Resolution & 56 & 112 & 224 & 448 & 56 & 112 & 224 & 448& 56 & 112 & 224 & 448\\
\midrule
DINO S/16 & 2.04 & 12.67 & 24.76 & 28.60 & 0.51 & 3.61  & 9.54  & 15.33 & 0.01 & 0.20 & 0.54 & 1.40 \\
DINO S/16 + LiFT  & 5.05 & 17.72 & 28.68 & 31.38 & 1.19 & 6.29  & 14.72 & 18.90 & 0.06 & 0.29 & 0.91 & 2.52 \\
\hdashline

DINO B/16   & 1.98  & 12.20 & 24.90 & 28.22 & 0.46 & 3.61  & 9.64  & 15.04 & 0.01 & 0.17 & 0.52 & 1.15 \\ 
DINO B/16 + LiFT  & 5.43 & 17.74 & 29.35 & 31.27 & 1.29 & 6.56  & 14.80 & 18.10 & 0.04 & 0.37 & 0.92 & 2.43 \\ 
\hdashline
DINO S/8  & 9.39  & 21.30 & 31.05 & 32.15 & 2.35 & 8.44  & 16.74 & 18.96 & 0.15 & 0.39 & 1.19 & 2.32 \\
DINO S/8 + LiFT	&12.90	&26.73	&34.54&	34.58	&4.35	&11.99	&20.61	&21.01&	0.18	&0.75&	2.21&	3.77 \\ 
\hdashline
DINO B/8  & 8.88  & 20.40 & 30.08 & 30.89 & 2.83 & 7.70  & 15.81 & 17.84 & 0.12 & 0.39 & 1.09 & 1.95 \\
DINO B/8 + LiFT & 12.21	&25.17	&33.23	&33.17	&4.22&	11.73	&19.39	&20.18&	0.13	&0.69&	2.27&	3.32\\ 
\hdashline
MOCO S/16	&1.27	&3.43	&7.37	&12.31&	0.21	&0.84	&2.35&	5.49&	0.00&	0.03	&0.10&	0.31\\
MOCO S/16 + LiFT&	6.48	&10.51	&14.13	&16.34	&1.74	&3.36	&6.42&	8.05	&0.04&	0.16&	0.42&	0.73\\
\hdashline

ViT S/16	&1.26&	5.72	&13.23	&16.90	&0.27	&1.62&	4.89&	7.34&	0.02	&0.06&	0.30&	0.50\\
ViT S/16 + LiFT	&3.76	&9.21&	16.58	&18.69	&1.02	&2.71	&6.63&	8.81&	0.02	&0.13	&0.45&	0.72\\

\hdashline
Leopart S/16 & 2.35  & 11.20 & 23.33 & 26.54 & 0.60 & 3.22  & 8.90  & 12.26 & 0.05 & 0.10 & 0.47 & 0.79 \\
Leopart S/16 + LiFT    & 4.24  & 15.61 & 27.77 & 30.06 & 1.22 & 5.16  & 12.81 & 15.66 & 0.02 & 0.25 & 0.74 & 1.39 \\ 
\bottomrule
\end{tabular}
}
\end{minipage}
\end{table*}

\section{Additional Details for ViTDet+LiFT}
\label{sec:apdx_vitdetlift}

For our experiments combining LiFT with ViTDet \cite{li2022exploring}, we increase the size of our LiFT module to address the additional complexity of the task and backbone. To be consistent with ViTDet, we use an MAE-trained ViT-Base backbone instead of the ViT-Small used in our other primary experiments.
Note that a standard ViT-Small model outputs feature maps with $384$ channels, while ViT-Base outputs $768$ channels. To handle the increased number of channels, we commensurately increase the number of channels in the layers of our LiFT module. We also add an additional convolutional block to the encoder segment. This larger LiFT module has a total of $7M$ parameters, as compared with the $1.2M$ parameter version used for smaller architectures. The ViTDet model used has $111M$ parameters, so our combined ViTDet+LiFT architecture has $118M$ parameters total. This is a $6.3\%$ increase in total parameters, which is similar to the relative percentage increase of the smaller LiFT version for DINO ViT-S/16.
Also, here we train LiFT on the COCO dataset in place of ImageNet. Because the COCO dataset is much smaller than ImageNet, we train on it for 100 epochs.

\section{Additional Backbones with LiFT}
\label{sec:apdx_backbones}

\begin{figure}[h]
    \centering
    \begin{subfigure}{0.9\linewidth}
        \centering
        \includegraphics[width=\linewidth]{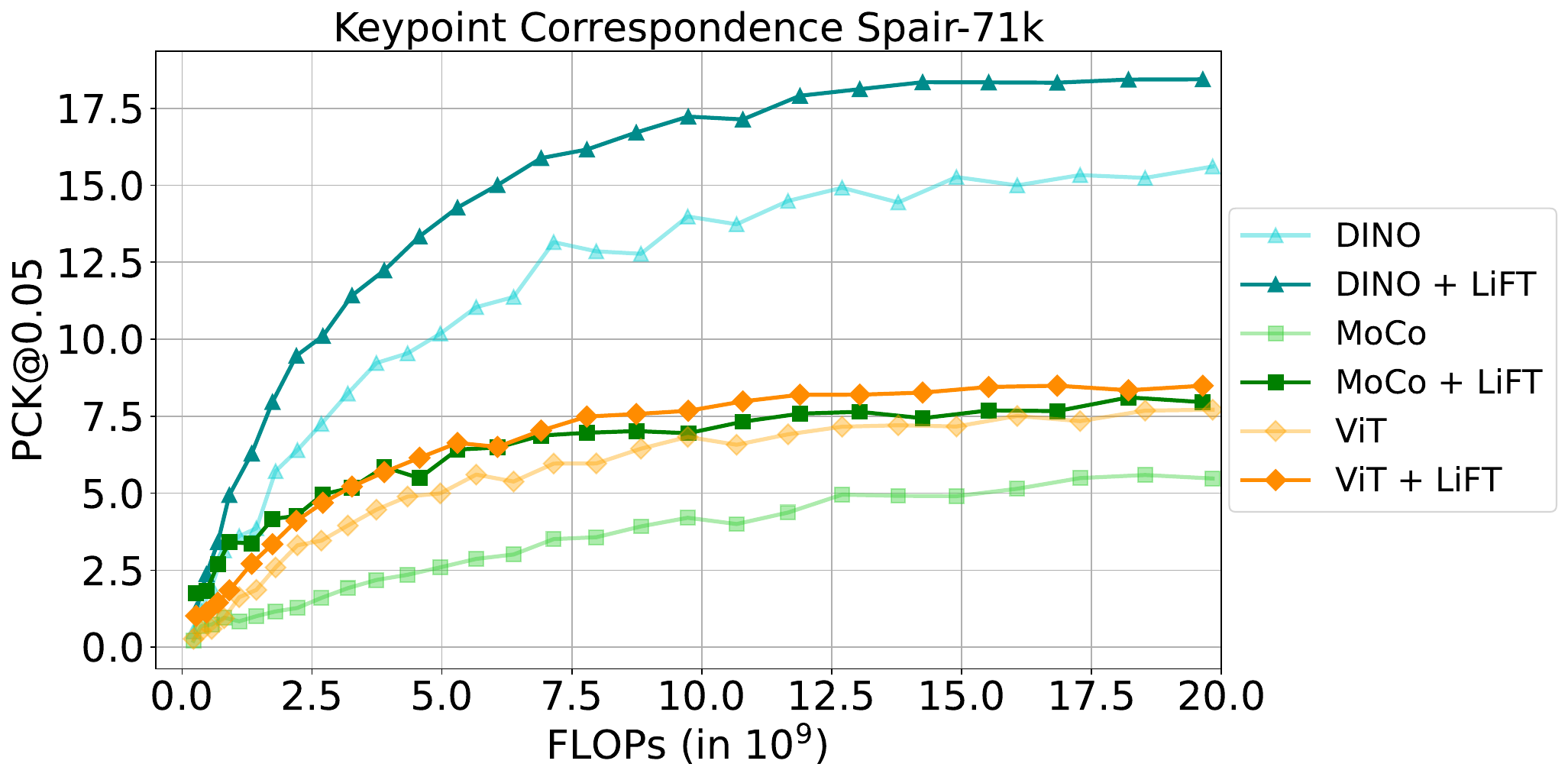}
    \end{subfigure}
    \caption{Performance \vs Compute Cost trade-off curve for LiFT when combined with different ViT backbones. Results are presented for SPair-71k Keypoint Correspondence. LiFT provides a performance boost for all three backbones at any FLOP budget.}
    \label{fig:backbone}
\end{figure}

We further demonstrate the general utility of LiFT by applying it to several additional backbones, including Leopart \cite{ziegler2022self} and several other DINO \cite{caron2021emerging} ViTs, namely ViT-S/8, ViT-B/16, and ViT-B/8. The results are summarized in Table \ref{tab:backbones_supp}. LiFT shows consistent improvement for the various architectures and models across patch sizes (8 and 16), trainings (Leopart and DINO), and backbone sizes (Base and Small).
We also extend the Performance \vs Compute Cost analysis curve to include both the MOCO and fully-supervised ViT backbones, as shown in Figure~\ref{fig:backbone}. We find that LiFT consistently boosts the performance of all three backbones at all FLOP allowances.

\begin{table}[t]
\begin{minipage}[!t]{\linewidth}
\setlength{\cmidrulewidth}{0.01em}
\renewcommand{\tabcolsep}{12pt}
\renewcommand{\arraystretch}{1.1}
\caption{Unsupervised Object Discovery comparison on PASCAL VOC 2007, PASCAL VOC 2012, and COCO20K. We report results for the CorLoc metric.}
\label{tab:supp_disc}
\centering
\resizebox{0.85\linewidth}{!}{
\begin{tabular}{@{}llcccc@{}}
\toprule
\multirow{2}{*}{Dataset} & \multirow{2}{*}{Method} & \multicolumn{4}{c}{Resolution} \\
\cmidrule[\cmidrulewidth](l){3-6}
& & 56 & 112 & 224 & 448 \\
\midrule
\multirow{7}{*}{VOC07} & DINO	&20.74&	50.07&	65.60&	68.27\\
& Leopart	&18.92	&32.59	&51.59	&48.79\\
& SelfPatch	&18.04	&41.99&	62.40	&63.62\\
& DINO+BL	& 21.27 & 46.96 & 64.70 & 68.37 \\
& DINO+RC	& 28.96 & 55.00 & 66.85 & 68.87 \\
& DINO+JBU	& 26.16 & 56.60 & 66.75 & 69.03 \\
& DINO+LiFT	&\textbf{36.54}	&\textbf{62.02}&	\textbf{68.79}	&\textbf{69.65}\\
\hdashline
\multirow{7}{*}{VOC12} & DINO&	23.27&	55.33	&69.01	&71.64\\
&Leopart	&22.44	&37.41&	55.74&	54.40\\
&SelfPatch&	20.19	&47.32&	68.02&	66.48\\
& DINO+BL	& 23.46 & 52.64 & 68.53 & 71.55 \\
& DINO+RC	& 31.63 & 59.96 & 68.87 & 71.47 \\
& DINO+JBU	& 28.97 & 61.67 & 69.13 & 71.54 \\
&DINO+LiFT&	\textbf{40.56}	&\textbf{66.21}&	\textbf{70.91}	&\textbf{71.71}\\
\hdashline
\multirow{7}{*}{COCO20K} & DINO	&16.28	&40.08&	53.98&	57.99\\
&Leopart	&16.14&	26.78	&43.89	&44.08\\
&SelfPatch	&14.15&	35.76&	52.18&	55.47\\
& DINO+BL	& 17.78 & 35.62 & 51.53 & 56.84 \\
&DINO+RC	& 22.92 & 42.53 & 54.52 & 58.40 \\
&DINO+JBU	& 21.36 & 43.87 & 55.45 & 58.82 \\
&DINO+LiFT	&\textbf{27.72}&	\textbf{50.20}&	\textbf{58.03}&	\textbf{60.50}\\
\bottomrule
\end{tabular}
}
\end{minipage}
\end{table}
\clearpage

\section{Additional Results}
\label{sec:apdx_metrics}

We present results for additional metrics on DAVIS, reporting the J Mean and F Mean in Table \ref{tab:davis_supp}. We present these results alongside the previously reported J \& F mean for completeness.
We find that both the J Mean and F Mean are also consistently improved by adding LiFT, and that DINO+LiFT surpasses all other baselines.
We also present additional Unsupervised Object Discovery results for PASCAL VOC 2007~\cite{pascal-voc-2007} and PASCAL VOC 2012~\cite{pascal-voc-2012} in Table \ref{tab:supp_disc} alongside the results for COCO20K. We again find that LiFT gives the best CorLoc boost over all baselines for both datasets.

\begin{table*}[t]
\begin{minipage}[t]{\linewidth}
\setlength{\cmidrulewidth}{0.01em}
\renewcommand{\tabcolsep}{6pt}
\renewcommand{\arraystretch}{1.1}
\caption{Comparison between LiFT and other baselines on the DAVIS Video Object Segmentation task with additional metrics J Mean and F Mean.}
\label{tab:davis_supp}
\resizebox{\linewidth}{!}{
\begin{tabular}{@{}lcccccccccccc@{}}
\toprule
& \multicolumn{4}{c}{J Mean} & \multicolumn{4}{c}{F Mean} & \multicolumn{4}{c}{J \& F Mean} \\
\cmidrule[\cmidrulewidth](l){2-5} \cmidrule[\cmidrulewidth](l){6-9} \cmidrule[\cmidrulewidth](l){10-13}
Method/Res. & 56 & 112 & 224 & 448 & 56 & 112 & 224 & 448& 56 & 112 & 224 & 448\\
\midrule
DINO&	9.50&	21.90&	37.80&	52.10&	5.20	&13.10&	28.10&	49.70	&7.40&	17.50&	33.00&	50.90\\
Leopart&	9.00	&20.20	&34.90	&47.30	&4.80	&12.10	&25.70	&42.80	&6.92&	16.12&	30.33&	45.08\\
SelfPatch&	9.70&	21.60&	38.10&	52.50	&5.10	&12.90	&27.90&	50.30&	7.40&	17.23&	33.01&	51.37\\
\hdashline
DINO+BL&	13.70&	29.40&	42.80	&54.00&	7.90	&17.90&	31.20&	52.00&	10.78&	23.66	&37.01	&53.02\\
DINO+RC&	13.80	&29.60&	43.00&	53.90	&8.20&	18.40	&31.90&	52.50	&11.00	&24.00&	37.40&	53.20\\
DINO+JBU&	13.90	&30.60&	42.80&	54.90	&8.60&	21.70&	35.10	&54.10&	11.20	&26.20&	39.00	&54.50\\
\hdashline
DINO+LiFT	&\textbf{16.27}&	\textbf{33.04}&	\textbf{48.07}	&\textbf{59.43}	&\textbf{9.72}&	\textbf{23.00}&	\textbf{40.56}	&\textbf{62.79}	&\textbf{13.00}	&\textbf{28.02}	&\textbf{44.32}&	\textbf{61.11}\\
\bottomrule
\end{tabular}
}
\end{minipage}
\end{table*}

\section{Additional Similarity Map Samples}
\label{sec:apdx_vis}

\begin{figure}[t!]
  \centering
    \includegraphics[width=0.55\linewidth]{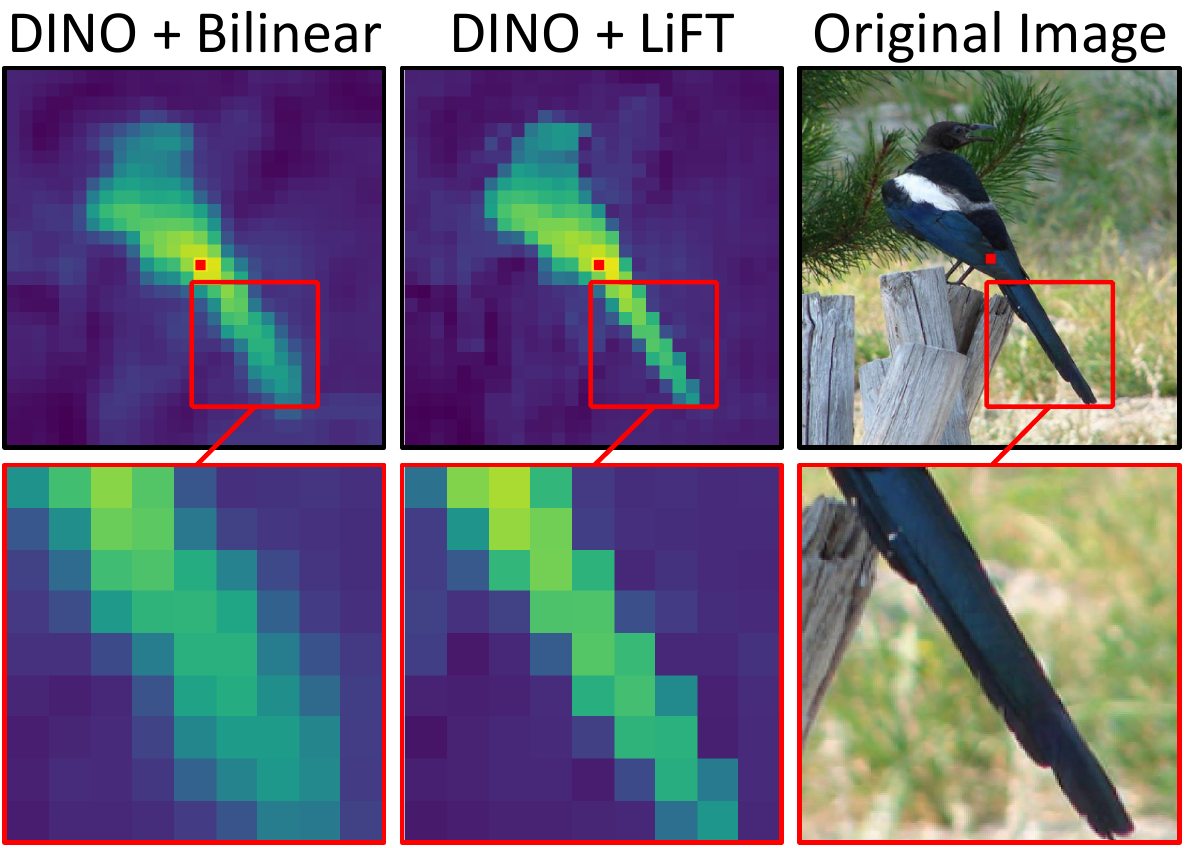}
    \caption{Compared with DINO+Bilinear, DINO+LiFT gives feature self-similarity maps with much sharper object boundaries, especially when zoomed in.}
    \label{fig:zoomviz}
\end{figure}

\begin{figure*}[h!]
  \centering
    \includegraphics[width=1.0\linewidth]{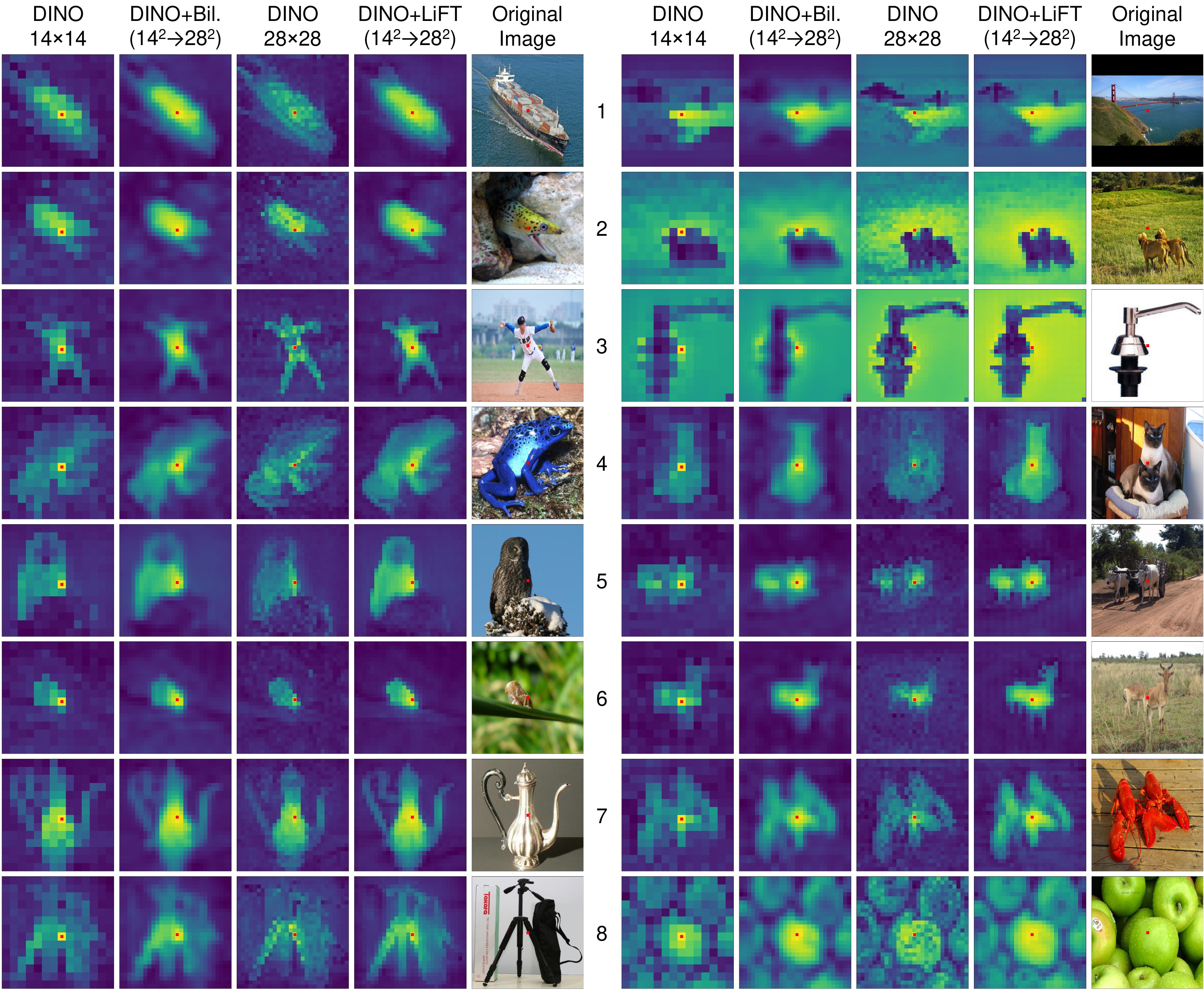}
    \caption{Additional visualizations of the self-similarity of features extracted from DINO, DINO+Bilinear interpolation, DINO with higher resolution image, and DINO+LiFT. The input image is shown for comparison. The self-similarity is computed using the feature corresponding to the center of the grid (marked in red) and all other features from each spatial location. Brighter pixels show a higher similarity.}
    \label{fig:supp_qualitative}
\end{figure*}

We have found that the feature self-similarity maps for DINO+LiFT more clearly and sharply outline the central object in an image. To further highlight this, we provide a zoomed-in comparison of the difference between DINO+LiFT and DINO+Bilinear upscaling in Figure \ref{fig:zoomviz}. We can see that DINO with Bilinear upsampling highlights the main object, but the outline is hazier and less precise due to the smoothing of the features. Meanwhile, the upscaled feature map produced by LiFT better respects object contours and produces a much sharper feature self-similarity map.
Finally, we provide additional samples further showing the benefits of LiFT for self-similarity maps, as shown in Figure~\ref{fig:supp_qualitative}. 
In rows 1 to 8 (left), we show samples with single central objects of differing shapes and sizes.
We see that the feature self-similarity maps for DINO+LiFT more uniformly fill the foreground object region, and have less noisy correlations with background regions.
In rows 1 to 3 (right), we show samples where the central feature vector, shown by the red marker,
lies on a background region. In these cases, we still see sharp contours around the foreground objects, or around the body of water in row 1 (right).
In cases like rows 4 to 8 (right), when there are multiple overlapping instances of the same object class, we see a uniform highlighting of the multiple object instances. We also see that DINO+LiFT better highlights thin structures in objects, like the teapot handle and tripod legs in rows 7 and 8 (left).
For comparison, when DINO (without LiFT) is given a doubled input size, these details are sometimes lost to noisy background regions.

\clearpage

\end{document}